%% file: main.tex
\documentclass[10pt,twocolumn,letterpaper]{article}

\usepackage[pagenumbers]{wacv} %

\input{preamble}

\definecolor{wacvblue}{rgb}{0.21,0.49,0.74}
\usepackage[pagebackref,breaklinks,colorlinks,allcolors=wacvblue]{hyperref}

\def\wacvPaperID{627} %
\def\confName{WACV}
\def\confYear{2027}

\title{\dataset{}: A Large-Scale Dataset and Benchmark for\\ \underline{M}ulti-\underline{M}odal \underline{M}ulti-session \underline{G}round-to-\underline{A}erial Place Recognition in Forests}

\author{
    Ethan Griffiths$^{1,2}$
    \quad{}Maryam Haghighat$^{1}$\quad{}Simon Denman$^{1}$\quad{}Clinton Fookes$^{1}$\quad{}Milad Ramezani$^{2}$\\
    $^1$Queensland University of Technology (QUT)\quad{}$^2$CSIRO Robotics \\
    $^1${\tt\small \{maryam.haghighat, s.denman, c.fookes\}@qut.edu.au
    }\\
    $^2${\tt\small \{ethan.griffiths, milad.ramezani\}@csiro.au\
    }
}

\begin{document}

\twocolumn[{%
\renewcommand\twocolumn[1][]{#1}%
\maketitle
\begin{center}
    \centering
    \captionsetup{type=figure}
    \includegraphics[width=0.97\textwidth]{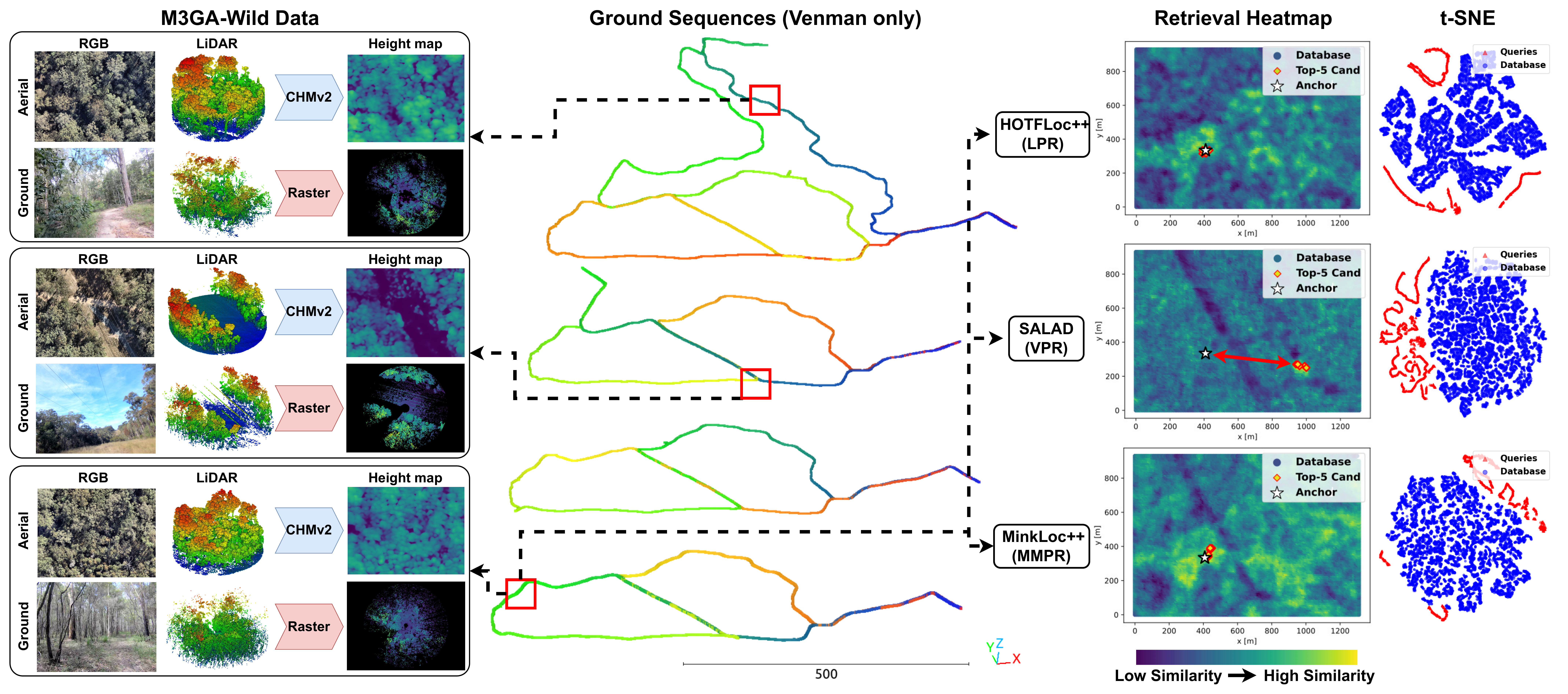}
    \vspace{-2 mm}
    \captionof{figure}{
    \dataset{}, a dataset and benchmark for multi-modal multi-session ground-to-aerial place recognition in forests. \emph{(Left)} Sample of several layers of data from one of four forests (Venman). \emph{(Middle)} Multi-session trajectories time-colourised indicating travel direction. \emph{(Right)} Retrieval heatmaps and t-SNE plots reveal perceptual aliasing and domain gaps in our dataset, particularly for vision-based methods.
    }
    \label{fig:hero}
\end{center}
}]
\input{sec/0_abstract}

\input{sec/1_intro}

\input{sec/2_relatedworks}

\input{sec/3_methodology}

\input{sec/4_experiments}

\input{sec/5_conclusion}

{
    \small
    \bibliographystyle{ieeenat_fullname}
    \bibliography{main}
}

\input{sec/X_supplementary}

\end{document}

%% file: preamble.tex
\usepackage{multirow}
\usepackage{makecell}       %
\usepackage{rotating}       %
\usepackage{siunitx}
\usepackage{balance}
\usepackage{pifont}%
\newcommand{\cmark}{\ding{51}}%
\newcommand{\xmark}{\ding{55}}%
\usepackage[accsupp]{axessibility}  %

\newcommand{\dataset}{M3GA-Wild}

\DeclareSIUnit{\pp}{p.p.}
\DeclareSIUnit{\nothing}{\relax}

%% file: sec/0_abstract.tex
\begin{abstract}

We present \dataset{}, the first benchmark for multi-modal, multi-session ground-to-aerial place recognition in forests. \dataset{} unifies and extends existing forest localisation datasets, providing a holistic benchmark with synchronised RGB imagery and LiDAR from ground traversals spanning 36 km, aligned high-resolution aerial imagery and multi-altitude LiDAR covering 370 hectares, and accurate geo-referenced 6-DoF poses for precise evaluation across four forests.
\dataset{} captures diverse forest scenes with varying viewpoints, occlusion, and environmental conditions, enabling systematic evaluation of visual, LiDAR, cross-modal, and multi-modal methods. Baseline experiments show that LiDAR-based approaches significantly outperform vision-only methods under severe viewpoint differences, while current multi-modal fusion strategies yield limited gains due to poor cross-modal alignment.
By pairing aerial RGB imagery with geo-referenced aerial LiDAR, \dataset{} also enables evaluation of foundation models for monocular depth estimation as a cheap source of 3D geometry from forest imagery, with initial experiments revealing shortfalls of current methods.
These results highlight key challenges in cross-platform localisation, including modality misalignment and severe domain gaps.
\dataset{} establishes a new benchmark to support research in robust multi-modal localisation and long-term autonomy in unstructured natural environments. The dataset and code is available at \url{https://github.com/csiro-robotics/M3GA-Wild}.
\vspace{-5mm}
\end{abstract}

%% file: sec/1_intro.tex
\section{Introduction}
\label{sec:intro}

Enabling robots to localise after deployment, or to ``wake up" under dense forest canopy without reliable GNSS is a core requirement for long-term distributed intelligence in natural settings.~Above-canopy data provides large-scale geo-referenced priors that can support the localisation of Unmanned Ground Vehicles (UGVs)~\cite{de2023air, de2025online}, allowing accurate collaborative operation in a unified coordinate system. 
However, identifying reliable place recognition (PR) cues perceivable from both ground and aerial viewpoints is more challenging than in urban areas.
Dense vegetation, repetitive geometric structure, seasonal appearance variation, uneven terrain, and illumination changes all degrade the reliability of conventional localisation~\cite{knightsWildPlacesLargeScaleDataset2023,knightsWildCrossCrossModalLarge2026}. 
High quality geo-referenced datasets with accurate 6-DoF ground-truth are thus essential for developing and benchmarking robust PR and localisation algorithms in these scenarios.
\begin{table*}[t]
\centering
\renewcommand{\arraystretch}{0.95}
\resizebox{0.95\textwidth}{!}{
\begin{tabular}{lcccccccc}
\toprule
\multirow{2.4}{*}{Dataset} & \multirow{2.4}{*}{Year} & \multicolumn{2}{c}{Data Modality} & \multicolumn{3}{c}{Viewpoints} & \multirow{2.4}{*}{Environment} & \multirow{2.4}{*}{Scale (Ground / Aerial)} \\
\cmidrule(lr){3-4} \cmidrule(lr){5-7}
 &  & RGB & LiDAR & Ground & LA & HA &  &  \\
\midrule

KITTI~\cite{geigerAreWeReady2012} & 2013 & \cmark & \cmark & \cmark & \xmark & \xmark & Urban & 44 km  \\

Oxford RobotCar~\cite{maddern1Year10002017} & 2017 & \cmark & \cmark & \cmark & \xmark & \xmark & Urban & 1000 km  \\

BotanicGarden~\cite{liuBotanicGardenHighQualityDataset2024} & 2020 & \cmark & \cmark & \cmark & \xmark & \xmark & Outdoor & 17 km  \\

MulRan~\cite{kimMulRanMultimodalRange2020} & 2020 & \xmark & \cmark & \cmark & \xmark & \xmark & Urban & 123 km  \\

Wild-Places~\cite{knightsWildPlacesLargeScaleDataset2023} & 2023 & \xmark & \cmark & \cmark & \xmark & \xmark & Forest & 33 km  \\

MARS-LVIG~\cite{liMARSLVIG2024} & 2024 & \cmark & \cmark & \xmark & \cmark & \xmark & Urban / Outdoor & 1.2 km$^2$  \\

SubT-MRS~\cite{zhaoSubTMRSDatasetPushing2024} & 2024 & \cmark & \cmark & \cmark & \xmark & \xmark & Sub-T / Various & 100 km  \\

Oxford Forest PR Dataset~\cite{ohEvaluationDeploymentLiDARbased2024} & 2024 & \xmark & \cmark & \cmark & \xmark & \xmark & Forest & 3.4 km  \\

WildCross~\cite{knightsWildCrossCrossModalLarge2026} & 2026 & \cmark & \cmark & \cmark & \xmark & \xmark & Forest & 33 km  \\

\midrule

CS-Campus3D~\cite{guanCrossLoc3DAerialGroundCrossSource2023} & 2023 & \xmark & \cmark & \cmark & \xmark & \cmark & Urban / Campus & 7.8 km / 5.5 km$^2$  \\

GRACO~\cite{zhuGRACOMultimodalDataset2023} & 2023 & \cmark & \cmark & \cmark & \cmark & \xmark & Campus & 3.0 km / 0.06 km$^2$  \\

CS-Wild-Places~\cite{griffithsHOTFormerLocHierarchicalOctree2025} & 2025 & \xmark & \cmark & \cmark & \cmark & \xmark & Forest & 36.1 km / 3.7 km$^2$  \\

\textbf{\dataset{} (Ours)} & 2026 & \cmark & \cmark & \cmark & \cmark & \cmark & Forest & 36.1 km / 3.7 km$^2$  \\

\bottomrule
\end{tabular}
}
\vspace{-2mm}
\caption{Comparison of PR and localisation datasets with LiDAR data. Many datasets focus on ground-to-ground PR. \dataset{} is the first to consider multi-modal ground-to-aerial place recognition in forests, with both low altitude (LA) and high altitude (HA) aerial data.}
\label{tab:dataset_comparison}
\vspace{-5mm}
\end{table*}

Recent studies have shown that leveraging cross-modal or multi-modal sensing data can improve localisation robustness in challenging environments~\cite{garcia-hernandezUnifyingLocalGlobal2024,gonzalezMultimodalLoopClosure2025,wangMultiModalAerialGroundCrossView2025}.
Vision-based approaches can be sensitive to illumination variation, motion blur, shadows, and dense foliage; while LiDAR-only systems may struggle in sparse, heavily occluded, or highly self-similar forest structures. The multi-modal combination of RGB appearance and LiDAR geometry provides complementary information that can improve robustness and generalisation across environmental conditions, sensor configurations, and viewpoints~\cite{komorowskiMinkLocLidarMonocular2021,laiAdaFusionVisualLiDARFusion2022}.~Existing PR datasets, however, either focus on urban driving, structured environments, or ground-level retrieval, or lack the combination of ground\,/\,aerial viewpoints, 6-DoF ground-truth, multi-session revisits, and calibrated multi-modal sensing needed to study cross-platform localisation in forests.

To address these limitations, this paper presents \dataset{}:~a novel large-scale \textbf{M}ulti-\textbf{M}odal \textbf{M}ulti-session benchmark for \textbf{G}round-to-\textbf{A}erial place recognition and localisation in forests~(\cref{fig:hero}). Building upon existing open-source benchmarks for ground-to-ground and ground-to-aerial forest PR, we extend the sensing coverage with high-resolution orthorectified aerial imagery and aerial LiDAR point clouds captured from high altitude. This enables comprehensive evaluation of ground-to-aerial PR and localisation across heterogeneous viewpoints, modalities, and overlap conditions. %
6-DoF ground-truth poses are generated via a geo-referenced LiDAR SLAM pipeline, enabling precise evaluation of cross-modal PR and multi-modal PR.

By combining challenging natural environments, multi-session revisits, heterogeneous sensor configurations, and varying viewpoints and degrees of spatial overlap, \dataset{} provides a dedicated benchmark for robust ground-to-aerial PR and multi-modal representation learning, enabling long-term autonomy in forests.
Beyond providing data and extensive evaluation protocols, \dataset{} is intended to shed light on the fundamental challenges of PR in dense forests, and serve as a reference benchmark for future research. By exposing the unique challenges of cross-platform localisation in forests, \dataset{} aims to demonstrate where existing methods fail and where new algorithmic advances are needed.

%% file: sec/2_relatedworks.tex
\section{Related Work}
\label{sec:related_works}
Place recognition has been widely studied across visual, LiDAR, cross-modal, and multi-modal settings.~Existing datasets and methods, however, remain largely centred on urban driving, structured man-made scenes, or from similar ground-level sensing, as summarised in~\cref{tab:dataset_comparison}.~While several datasets support visual PR (VPR)~\cite{ali2022gsv, weyand2020google}, LiDAR PR (LPR)~\cite{knightsWildPlacesLargeScaleDataset2023, oh2024evaluation}, cross-modal PR (CMPR) or multi-modal PR (MMPR)~\cite{maddern1Year10002017, knightsWildCrossCrossModalLarge2026}, few address ground-to-aerial PR in forests, particularly where RGB and LiDAR are available from both viewpoints. This leaves an important gap for evaluating ground localisation systems intended for GPS-degraded, unstructured and vegetation-rich environments while leveraging increasingly accessible geo-referenced aerial LiDAR and imagery as prior maps.

In urban areas, PR performance has improved substantially and, in some settings, is approaching saturation as reflected by results from the General Place Recognition Challenge at IROS 2023\footnote{\url{https://github.com/MetaSLAM/GPR_Competition}\label{fn:gcpr}}. Methods trained on one urban dataset often transfer well to unseen urban environments~\cite{keethaAnyLocUniversalVisual2024,vidanapathiranaSpectralGeometricVerification2023}, as RGB and LiDAR contain distinctive and persistent cues from buildings, road layouts and other man-made structures. These cues provide strong geometric and semantic anchors that neural networks can use, making even ground-to-aerial matching more tractable in cities  ~\cite{zaffarAreStateoftheartVisual2019, jungHeLiOSHeterogeneousLiDAR2025}. 

Unlike urban areas, forest scenes lack such stable man-made landmarks and are dominated by repeated, self-similar structures such as trunks, branches, foliage, and canopy layers. As a result, PR cues that are distinctive and repeatable in cities often become ambiguous or unreliable in forests. In the following, we briefly review VPR, LPR, CMPR and MMPR methods, many of which have been developed or evaluated in urban areas, and discuss in~\cref{sec:experiment} the challenges of transferring them to forest settings.

VPR methods primarily focus on learning compact and discriminative image descriptors~\cite{arandjelovicNetVLADCNNArchitecture2016,ali-beyMixVPRFeatureMixing2023,izquierdoOptimalTransportAggregation2024,ali-beyBoQPlaceWorth2024}, whilst recent methods leverage foundation model features for improved generalisation.
These methods perform well in urban and road environments where persistent landmarks provide distinctive visual cues, but image-only retrieval becomes substantially more ambiguous in forests. This limitation is reflected in the General Place Recognition Challenge at ICRA 2022 based on Visual Terrain Relative Navigation Dataset\footref{fn:gcpr} which includes forest and rural scenes but evaluates only aerial-to-aerial retrieval, leaving the more severe viewpoint and visibility gaps of ground-to-aerial PR unexplored.

LPR addresses some limitations of VPR methods by exploiting geometric structure rather than appearance. Recent methods leverage bird's-eye-view (BEV) representations~\cite{shenForestLPRLiDARPlace2025}, sparse CNNs~\cite{komorowskiMinkLoc3DPointCloud2021, vidanapathiranaLoGG3DNetLocallyGuided2022}, or transformers~\cite{goswamiSALSASwiftAdaptive2024,griffithsHOTFLocEndtoEndHierarchical2025, griffithsHOTFormerLocHierarchicalOctree2025} for descriptor extraction. While generally more robust to illumination and texture changes, most LPR methods are developed for same-platform or ground-to-ground retrieval.
Recent methods such as HOTFormerLoc~\cite{griffithsHOTFormerLocHierarchicalOctree2025} and HOTFLoc++~\cite{griffithsHOTFLocEndtoEndHierarchical2025} have begun addressing ground-to-aerial localisation in forests, but remain restricted to LiDAR-only PR.

CMPR or MMPR methods bridge modality gaps by combining visual and geometric cues.~CMPR enables deployment of heterogeneous platforms in resource-constrained settings, while MMPR leverages multiple sensors for improved robustness.~RGB2LIDAR matches ground RGB images to aerial LiDAR-derived depth maps~\cite{mithunRGB2LIDARSolvingLargeScale2020}, while LIP-Loc learns a shared image--point cloud embedding space~\cite{puligillaLIPLocLiDARImage2024}.~Works such as AGI2P~\cite{yangAGI2PBenchmarkingAerial2026} and ground-to-aerial image localisation methods~\cite{ferversUncertaintyAwareVisionBasedMetric2023,fervers2024statewide} explore cross-view matching, while MinkLoc++~\cite{komorowskiMinkLocLidarMonocular2021}, AdaFusion~\cite{laiAdaFusionVisualLiDARFusion2022}, spherical early fusion ~\cite{bernreiter2021spherical}, and UMF~\cite{garcia-hernandezUnifyingLocalGlobal2024} combine RGB and LiDAR through different fusion strategies. However, evaluations are largely limited to less challenging urban, semi-structured, or same-platform settings.

In addition, advances in feed-forward geometric foundation models~\cite{wangDepthAnythingAny2025, brandtCHMv2ImprovementsGlobal2026} have made dense monocular depth prediction increasingly effective, creating new opportunities to study visual-geometric representations in natural environments~\cite{knightsWildCrossCrossModalLarge2026}. However, depth prediction in forests remains under-explored, particularly for aerial imagery where canopy structure, shadows, vegetation density, and height ambiguity challenge metric estimation. By providing paired aerial RGB and LiDAR data, our dataset enables predicted depth to be evaluated as ``pseudo'' aerial LiDAR for PR, testing whether image-derived geometry can reduce reliance on aerial LiDAR for robust ground-to-aerial retrieval.

%% file: sec/3_methodology.tex
\section{\dataset{} Dataset}
\label{sec:method}

This section describes the design, sensing configuration, data generation, and benchmarking protocol of \dataset{}, particularly to support research into multi-modal and cross-modal ground-to-aerial localisation under canopies.

\dataset{} is organised as a unified multi-layer benchmark in which synchronised ground image and LiDAR traversals are aligned with low altitude (LA) LiDAR, high altitude (HA) LiDAR, aerial RGB imagery, and derived height representations. We define LA and HA as aerial LiDAR captured between \qtyrange{50}{200}{\m} and \qtyrange{800}{2500}{\m} above ground, respectively.
Building upon two existing datasets, we combine ground LiDAR submaps and RGB imagery from WildCross~\cite{knightsWildCrossCrossModalLarge2026} with additional ground LiDAR sequences and corresponding geo-referenced LA LiDAR submaps from CS-WildPlaces~\cite{griffithsHOTFormerLocHierarchicalOctree2025}, covering four distinct eucalyptus forests in Brisbane, Australia: Karawatha, Venman, Samford, and QCAT. On top of this rich data, \dataset{} introduces an aerial RGB imagery layer aligned with the LA LiDAR data, and geo-referenced HA LiDAR scans, providing a lower resolution but scalable alternative to drone LiDAR scans. A summary of \dataset{} sequences can be found in the supplementary.

We derive secondary data products from the primary sensing streams to investigate if alternative representations reduce the ground--aerial domain gap. These include ground LiDAR-derived height maps, and canopy height maps predicted from aerial imagery using recent DINOv3-based foundation models. The result is a holistic dataset that enables benchmarking of PR and localisation across heterogeneous sensing modalities and viewpoints in dense forests.

\subsection{Data Collection}
\label{sec:method_data_collection}

\noindent\textbf{Ground 2D\,/\,3D Data:}\quad
We use the ground data from WildCross and CS-Wild-Places, collected via handheld perception pack equipped with forward-facing RGB camera\footnote{Ground RGB data is not currently available for QCAT and Samford.\label{fn:gnd_rgb}} and spinning VLP-16 LiDAR. LiDAR-inertial SLAM~\cite{ramezaniWildcatOnlineContinuousTime2022} was employed to generate globally consistent maps and near-ground truth trajectories, with GNSS, IMU, and LiDAR integration. We refer readers to WildCross~\cite{knightsWildCrossCrossModalLarge2026} for further details on Karawatha and Venman data collection, and CS-Wild-Places~\cite{griffithsHOTFormerLocHierarchicalOctree2025} for details on QCAT and Samford. 

\noindent\textbf{Aerial 2D\,/\,3D Data:}\quad
Two drone configurations were used for low altitude LiDAR data collection. For Karawatha, Venman, and QCAT, a DJI M300 quadcopter with VLP-32C LiDAR sensor was deployed. For Samford, an Acecore NOA hexacopter equipped with RIEGL VUX-120 LiDAR was used, providing higher point density and canopy penetration. Both drones flew in a lawnmower pattern over each forest. GNSS RTK ensured precise geo-registration, and overlapping ground and aerial scans were aligned using ICP until RMSE between correspondences was $\leq$ \qty{0.5}{\m}. We refer readers to~\cite{griffithsHOTFormerLocHierarchicalOctree2025} for further details.

To build a paired aerial RGB--LiDAR database, we utilise and curate high-resolution Nearmap\footnote{\url{https://www.nearmap.com/au/products/imagery}\label{fn:nearmap}} imagery, aligning it to the drone LiDAR submap grids. The RTK geo-referenced imagery is captured within two months of each aerial LiDAR scan from an aircraft equipped with a nadir-facing camera, with up to \qty{7.5}{\cm} ground sample distance (GSD) and \qty{25.5}{\cm} RMSE horizontal accuracy.

\noindent\textbf{High Altitude LiDAR:}\quad
We introduce a novel high altitude (HA) layer of LiDAR data to our dataset. HA LiDAR is a cheaper, more accessible alternative to LA LiDAR; capable of scanning large areas in a single flight.
However, it typically achieves lower point density and vegetation penetration due to increased flight height. By providing both sources of LiDAR data, \dataset{} enables insights into which is required for different tasks. A visual comparison of each LiDAR source is available in the supplementary.

Specifically, we use publicly-available LiDAR data captured by the Queensland Government between 2018 to 2022~\cite{thestateofqueenslandQueenslandLiDARData2019}, with 2-$\sigma$ horizontal and vertical accuracy of \qty{80}{\cm} and \qty{30}{\cm}, respectively. Each scan is geo-registered with GNSS RTK, and ICP aligned to aerial data following the approach used for LA LiDAR. Similar to drone LiDAR data, we combine this layer with aligned Nearmap\footref{fn:nearmap} aerial RGB imagery captured within two months of each scan.

\subsection{Pre-Processing}
Pre-processing converts raw data into a format suitable for multi-modal PR evaluation.~\,Ground and aerial LiDAR scans are processed into submaps following the approach in CS-Wild-Places~\cite{griffithsHOTFormerLocHierarchicalOctree2025}, producing over \qty{65}{\kilo\nothing} ground submaps, and over \qty{70}{\kilo\nothing} aerial submaps spanning a \qty{10}{\m}-spaced grid covering each forest, each cropped to a \qty{30}{\m} horizontal radius. For training deep PR networks, we remove ground points with a Cloth Simulation Filter~\citep{zhangEasytoUseAirborneLiDAR2016} and downsample with a \qty{0.8}{\m} voxel grid filter. Additionally, we include point-wise intensity information, and save submaps in the KITTI~\cite{geigerAreWeReady2012} binary format. Ground RGB images follow WildCross~\cite{knightsWildCrossCrossModalLarge2026} pre-processing, producing over \qty{376}{\kilo\nothing} rectified front-facing images at \qty{15}{\hertz}.
Aerial imagery is cropped to patches aligned with aerial LiDAR submaps, such that the centre of each image corresponds to the grid coordinate of each submap centroid, with ground coverage equivalent to \qty{83}{\m}$\times$\qty{63}{\m}. For ease of use, we downsample aerial images to the same resolution and aspect ratio as the downsampled ground images ($504\,\text{px}\times378$\,px).

\subsection{Training and Testing Splits}
\label{sec:train_test_splits}

We follow the training and evaluation protocol of~\cite{griffithsHOTFormerLocHierarchicalOctree2025}, training on ground and aerial sequences of Karawatha and Venman, with spatially disjoint samples withheld for testing. To avoid leakage, all training samples overlapping the test set are excluded, and the K-01 test set is used for validation. QCAT and Samford serve as unseen test sets for evaluating domain-shift with all modalities but ground images. During training, we consider a \qty{15}{\m} positive and \qty{60}{\m} negative horizontal threshold between ground queries and potential aerial pairs, and a \qty{30}{\m} threshold during testing.~\Cref{tab:train_test_stats} summarises the training, validation, and testing sets, with visualisations of each provided in the supplementary.

\begin{table}[t]
    \centering
    \renewcommand{\arraystretch}{0.95}
    \setlength{\tabcolsep}{2.5pt} %

    \resizebox{\linewidth}{!}{
    \normalsize
    \begin{tabular}{lcccccccc}
    \toprule
    \multirow{3.8}{*}{Forest} & \multicolumn{4}{c}{Images} & \multicolumn{4}{c}{LiDAR Submaps} \\
    \cmidrule(lr){2-5} \cmidrule(lr){6-9}
    & \multirow{2.3}{*}{Train} & \multirow{2.3}{*}{Val} & \multicolumn{2}{c}{Test} & \multirow{2.3}{*}{Train} & \multirow{2.3}{*}{Val} & \multicolumn{2}{c}{Test} \\
    \cmidrule(lr){4-5} \cmidrule(lr){8-9}
    & & & Query & DB & & & Query & DB \\
    \midrule
    Karawatha & 231.8\,k & 6.8\,k & 11.2\,k & 17.8\,k & 37.3\,k & 2.5\,k & 7.4\,k & 17.8\,k \\
    Venman    & 124.4\,k & --       & 9.6\,k  & 12.4\,k & 26.7\,k & --      & 6.4\,k & 12.4\,k \\
    QCAT      & --       & --       & --      & 0.4\,k  & --      & --      & 0.8\,k & 0.4\,k \\
    Samford   & --       & --       & --      & 4.6\,k  & --      & --      & 1.3\,k & 4.6\,k \\
    \midrule
    Total     & 356.2\,k & 6.8\,k & 20.9\,k & 35.1\,k & 64.0\,k & 2.5\,k & 15.9\,k & 35.1\,k \\
    \bottomrule
    \end{tabular}
    }
    \vspace{-.2 cm}
    \caption{Dataset statistics for each forest region of \dataset{}.}
    \label{tab:train_test_stats}
    \vspace{-5mm}
\end{table}

\begin{table*}[t]
\centering
\renewcommand{\arraystretch}{0.95}
\resizebox{\linewidth}{!}{
\begin{tabular}{llccccccccccccccccccc}
\toprule
& \multirow{2}{*}{Method} & \multirow{2}{*}{Re-Ranker} & \multicolumn{3}{c}{Karawatha} & \multicolumn{3}{c}{Venman} & \multicolumn{3}{c}{QCAT} & \multicolumn{3}{c}{Samford} & \multicolumn{3}{c}{Average} \\
& & & R1 & R5 & R10 & R1 & R5 & R10 & R1 & R5 & R10 & R1 & R5 & R10 & R1 & R5 & R10  \\
\midrule
\multirow{9.5}{*}{\rotatebox{90}{Low Altitude (LA)}} & MinkLoc3Dv2~\cite{komorowskiImprovingPointCloud2022} & --- & 46.2 & 63.0 & 70.1 & \textbf{63.6} & \textbf{82.7} & \textbf{90.5} & 49.5 & 73.0 & 83.8 & 34.4 & 53.1 & 61.2 & 48.5 & 67.9 & 76.4 \\
& EgoNN~\cite{komorowskiEgoNNEgocentricNeural2022} & --- & 45.4 & 63.4 & 73.1 & 56.4 & 78.8 & 86.7 & 42.1 & 75.4 & 85.8 & 49.3 & 68.2 & 76.8 & 48.3 & 71.4 & 80.6 \\
& LoGG3D-Net~\cite{vidanapathiranaLoGG3DNetLocallyGuided2022} & --- & 21.7 & 37.2 & 45.5 & 27.7 & 43.1 & 52.1 & 31.1 & 50.3 & 56.8 & 7.3 & 12.2 & 15.5 & 21.9 & 35.7 & 42.5 \\
& CrossLoc3D~\cite{guanCrossLoc3DAerialGroundCrossSource2023} & --- & 7.9 & 20.0 & 26.4 & 11.7 & 28.1 & 38.6 & 16.5 & 37.9 & 55.8 & 4.0 & 11.7 & 16.0 & 10.0 & 24.4 & 34.2 \\
& HOTFormerLoc~\cite{griffithsHOTFormerLocHierarchicalOctree2025} & --- & 50.2 & 65.5 & 72.7 & 50.8 & 67.8 & 76.1 & \textbf{62.8} & 83.1 & \textbf{92.3} & 73.8 & 85.1 & 90.3 & 59.4 & 75.4 & 82.8 \\
& HOTFLoc++~\cite{griffithsHOTFLocEndtoEndHierarchical2025} & --- & \textbf{63.2} & \textbf{75.9} & \textbf{81.9} & 56.3 & 72.4 & 80.6 & 61.1 & \textbf{83.5} & 90.9 & \textbf{88.8} & \textbf{95.5} & \textbf{97.1} & \textbf{67.4} & \textbf{81.8} & \textbf{87.6} \\
\cmidrule{2-18}
& EgoNN~\cite{komorowskiEgoNNEgocentricNeural2022} & SGV~\cite{vidanapathiranaSpectralGeometricVerification2023} & 64.7 & 73.5 & 77.0 & 67.7 & 84.4 & 89.4 & 66.6 & 81.9 & 87.5 & 79.8 & 81.9 & 83.0 & 69.7 & 80.4 & 84.3 \\
& LoGG3D-Net~\cite{vidanapathiranaLoGG3DNetLocallyGuided2022} & SGV~\cite{vidanapathiranaSpectralGeometricVerification2023} & 50.1 & 52.2 & 53.4 & 60.1 & 61.7 & 62.4 & 52.0 & 59.2 & 62.2 & 10.7 & 12.8 & 14.3 & 43.2 & 46.5 & 48.1 \\
& HOTFLoc++~\cite{griffithsHOTFLocEndtoEndHierarchical2025} & MSGV~\cite{griffithsHOTFLocEndtoEndHierarchical2025} & \textbf{78.3} & \textbf{84.2} & \textbf{86.1} & \textbf{86.6} & \textbf{87.8} & \textbf{88.1} & \textbf{91.3} & \textbf{95.1} & \textbf{95.7} & \textbf{95.8} & \textbf{97.7} & \textbf{98.3} & \textbf{88.0} & \textbf{91.2} & \textbf{92.0} \\
\midrule
\multirow{8.4}{*}{\rotatebox{90}{High Altitude (HA)}} & MinkLoc3Dv2~\cite{komorowskiImprovingPointCloud2022} & --- & 66.9 & 82.3 & 87.5 & 69.8 & 87.2 & 92.0 & 84.8 & 96.5 & 98.7 & 23.8 & 38.5 & 46.3 & 61.3 & 76.1 & 81.1 \\
& EgoNN~\cite{komorowskiEgoNNEgocentricNeural2022} & --- & 58.9 & 73.0 & 78.9 & 70.4 & \textbf{91.2} & \textbf{96.2} & \textbf{93.2} & 97.9 & \textbf{99.0} & 43.7 & 62.2 & 69.3 & 66.5 & 81.1 & 85.8 \\
& LoGG3D-Net~\cite{vidanapathiranaLoGG3DNetLocallyGuided2022} & --- & 25.9 & 41.4 & 46.9 & 25.5 & 47.4 & 57.6 & 58.8 & 82.1 & 90.3 & 5.9 & 14.4 & 22.3 & 29.0 & 46.3 & 54.3 \\
& HOTFormerLoc~\cite{griffithsHOTFormerLocHierarchicalOctree2025} & --- & 67.5 & 81.6 & 86.3 & 55.6 & 71.9 & 77.8 & 88.2 & 97.1 & 98.3 & 48.4 & 63.0 & 69.4 & 64.9 & 78.4 & 83.0 \\
& HOTFLoc++~\cite{griffithsHOTFLocEndtoEndHierarchical2025} & --- & \textbf{67.9} & \textbf{83.0} & \textbf{88.6} & \textbf{74.3} & 88.5 & 93.6 & 92.6 & \textbf{98.4} & \textbf{99.0} & \textbf{49.9} & \textbf{67.5} & \textbf{74.4} & \textbf{71.2} & \textbf{84.4} & \textbf{88.9} \\  %
\cmidrule{2-18}
& EgoNN~\cite{komorowskiEgoNNEgocentricNeural2022} & SGV~\cite{vidanapathiranaSpectralGeometricVerification2023} & 72.4 & 79.5 & 82.0 & 85.3 & 95.2 & \textbf{97.5} & \textbf{97.9} & \textbf{99.4} & \textbf{99.4} & 50.6 & 69.3 & 74.4 & 76.6 & 85.8 & 88.3 \\
& LoGG3D-Net~\cite{vidanapathiranaLoGG3DNetLocallyGuided2022} & SGV~\cite{vidanapathiranaSpectralGeometricVerification2023} & 49.4 & 50.7 & 51.1 & 67.6 & 68.0 & 68.2 & 95.2 & 96.5 & 96.8 & 22.8 & 26.8 & 29.9 & 58.8 & 60.5 & 61.5 \\
& HOTFLoc++~\cite{griffithsHOTFLocEndtoEndHierarchical2025} & MSGV~\cite{griffithsHOTFLocEndtoEndHierarchical2025} & \textbf{87.3} & \textbf{91.6} & \textbf{92.7} & \textbf{96.2} & \textbf{97.1} & 97.4 & 97.5 & 98.7 & 99.0 & \textbf{72.5} & \textbf{76.3} & \textbf{78.6} & \textbf{88.4} & \textbf{90.9} & \textbf{91.9} \\ %
\bottomrule
\end{tabular}
}
\vspace{-.2 cm}
\caption{LiDAR place recognition results on \dataset{}, for both low and high altitude LiDAR data.}
\label{tab:lpr}
\vspace{-.5 cm}
\end{table*}

\subsection{Secondary Products}

We explore how foundation models for monocular depth estimation (MDE) transfer to forest environments. In particular, we leverage CHMv2~\cite{brandtCHMv2ImprovementsGlobal2026}; a DINOv3-based foundation model for forest canopy height estimation trained on global satellite imagery. We process all aerial imagery in \dataset{} with CHMv2 and provide predicted height maps as an additional layer of pseudo-3D data, opening new avenues for CMPR using 3D structure inferred from cheaply available aerial imagery. See \cref{fig:hero} for comparisons of CHMv2 outputs. As CHMv2 was originally trained on \qty{1}{\m} resolution satellite imagery, it struggles to estimate fine-grained height variations from our high resolution images without fine-tuning. We also tested Prior Depth Anything~\cite{wangDepthAnythingAny2025} with high altitude aerial LiDAR as a sparse prior, but it failed to produce reasonable height estimates from aerial imagery.

%% file: sec/4_experiments.tex
\section{Experiments}
\label{sec:experiment}
We evaluate VPR, LPR, CMPR, and MMPR methods on \dataset{}, reporting mean Recall@1\,/\,5\,/\,10 (R1, R5, R10).~In all experiments, ground samples form queries against the aerial database for each region, following \cref{sec:train_test_splits} (low altitude LiDAR is used unless otherwise specified). 
We further analyse which modality combinations improve cross-view alignment, how RGB and LiDAR contribute to PR, and how PR performance changes when a coarse pseudo-GNSS prior narrows the aerial search space. Finally, we summarise experimental insights and provide practical recommendations for deploying such PR systems in forests. See implementation details in the Supplementary.

\subsection{LiDAR Place Recognition}
\label{sec:lpr}
We evaluate LPR approaches on \dataset{} under two settings: low altitude (LA) and high altitude (HA).
Models are trained on either LiDAR source to compare the effects of flight height on performance.
Additionally, we compare re-ranking methods which analyse local features of the top-20 candidates to re-order retrievals based on geometric fitness. 

Across LPR methods (\cref{tab:lpr}), we observe strong performance despite the severe ground-to-aerial domain gap. This highlights the advantage of LiDAR geometry in forests, even under strong occlusions. HOTFLoc++~\cite{griffithsHOTFLocEndtoEndHierarchical2025} achieves the best mean performance on LA LiDAR, reaching 67.4\%\,/\,81.8\%\,/\,87.6\% average R1\,/\,R5\,/\,R10, outperforming popular learnable methods such as MinkLoc3Dv2~\cite{komorowskiImprovingPointCloud2022} (48.5\%\,/\,67.9\%\,/\,76.4\%) and LoGG3D-Net~\cite{vidanapathiranaLoGG3DNetLocallyGuided2022} (21.9\%\,/\,35.7\%\,/\,42.5\%). With re-ranking, it achieves 88.0\%\,/\,91.2\%\,/\,92.0\% on average with multi-scale geometric verification (MSGV): an R1 improvement of 18.3 percentage points over EgoNN~\cite{komorowskiEgoNNEgocentricNeural2022} with SpectralGV~\cite{vidanapathiranaSpectralGeometricVerification2023}.

Performance on HA LiDAR is better than LA LiDAR for most methods on all forests except Samford. HOTFLoc++ achieves the best performance, reaching 88.4\%\,/\,90.9\%\,/\,91.9\% average R1\,/\,R5\,/\,R10 with MSGV. This seems counter-intuitive at first, as HA LiDAR has reduced point density and canopy penetration compared with LA LiDAR. However, coarse voxelisation used by each method reduces the advantage of high point density. Additionally, HA LiDAR scans are more consistent due to the near-identical setup used for all flights. By contrast, different drone, flight, and LiDAR configurations were used for each of the LA scans, introducing differences in point distribution that affects the transferability of learned features.

Across all LPR experiments, transformer-based hierarchical encoders maintain the greatest robustness to different environments and data types. This suggests that modelling long-range spatial context is critical in environments with sparse and repetitive geometric structure, particularly for the low overlap conditions of ground-to-aerial settings.

\begin{table}[t]
\centering
\renewcommand{\arraystretch}{0.95}
\resizebox{\linewidth}{!}{
\begin{tabular}{llccccccccc}
\toprule
& \multirow{2}{*}{Method} & \multicolumn{3}{c}{Karawatha} & \multicolumn{3}{c}{Venman} & \multicolumn{3}{c}{Average} \\
& & R1 & R5 & R10 & R1 & R5 & R10 & R1 & R5 & R10  \\
\midrule
\multirow{6}{*}{\rotatebox{90}{Zero-shot}} & NetVLAD~\cite{arandjelovicNetVLADCNNArchitecture2016} & 0.40 & 1.01 & 1.79 & 0.27 & 1.24 & 2.39 & 0.33 & 1.12 & 2.09 \\
& MixVPR~\cite{ali-beyMixVPRFeatureMixing2023} & 0.60 & 1.92 & 3.23 & 0.14 & 0.49 & 1.14 & 0.37 & 1.21 & 2.18 \\
& SALAD~\cite{izquierdoOptimalTransportAggregation2024} & 0.54 & 1.64 & 2.92 & 0.25 & 0.73 & 1.42 & 0.40 & 1.18 & 2.17 \\
& BoQ~\cite{ali-beyBoQPlaceWorth2024} & 0.92 & 2.67 & 3.85 & 0.08 & 0.36 & 0.79 & 0.50 & 1.51 & 2.32 \\
& AnyLoc~\cite{keethaAnyLocUniversalVisual2024} & \textbf{2.11} & \textbf{4.43} & \textbf{6.10} & \textbf{0.72} & \textbf{2.51} & 4.52 & \textbf{1.41} & \textbf{3.47} & \textbf{5.31} \\
& AnyLoc$^\dagger$~\cite{keethaAnyLocUniversalVisual2024} & 0.50 & 1.94 & 2.87 & 0.35 & 2.19 & \textbf{4.88} & 0.43 & 2.07 & 3.88 \\
\midrule
\multirow{4}{*}{\rotatebox{90}{Fine-tuned}} & NetVLAD~\cite{arandjelovicNetVLADCNNArchitecture2016} & 0.17 & 0.56 & 0.87 & \textbf{0.51} & \textbf{1.65} & \textbf{2.80} & 0.34 & 1.11 & 1.84 \\
& MiXVPR~\cite{ali-beyMixVPRFeatureMixing2023} & 0.26 & 1.07 & 2.22 & 0.30 & 0.79 & 1.30 & 0.28 & 0.93 & 1.76 \\
& SALAD~\cite{izquierdoOptimalTransportAggregation2024} & 0.97 & 2.85 & 4.66 & 0.29 & 0.97 & 2.66 & 0.63 & 1.91 & \textbf{3.66} \\
& BoQ~\cite{ali-beyBoQPlaceWorth2024} & \textbf{3.37} & \textbf{4.70} & \textbf{5.80} & 0.09 & 0.47 & 1.39 & \textbf{1.73} & \textbf{2.58} & 3.59 \\
\bottomrule
\multicolumn{11}{l}{$^\dagger$ indicates the `unstructured' domain is used for AnyLoc cluster centres.}
\end{tabular}
}
\vspace{-3mm}
\caption{Visual place recognition results on \dataset{}.}
\label{tab:vpr}
\vspace{-6mm}
\end{table}

\subsection{Visual Place Recognition}
\label{sec:vpr}
To evaluate VPR, we report results for two settings: Zero-shot and Fine-tuned. This quantifies the cross-domain gap for VPR models trained on urban environments, examining their adaptation to ground-aerial forest settings.

VPR methods (\cref{tab:vpr}) exhibit extremely low absolute performance in the zero-shot setting. This is somewhat expected given the extreme domain gap between ground-level imagery (largely occluded by vegetation) and aerial views (top-down canopy texture with limited structural cues).
Even strong modern visual descriptors such as MixVPR~\cite{ali-beyMixVPRFeatureMixing2023}, SALAD~\cite{izquierdoOptimalTransportAggregation2024}, and BoQ~\cite{ali-beyBoQPlaceWorth2024} fail to generalise effectively, with R1 typically below 1\%. The best-performing zero-shot method, AnyLoc~\cite{keethaAnyLocUniversalVisual2024}, reaches only 1.41\% R1, indicating that even foundation-model features struggle to bridge the extreme cross-view mismatch in dense forest environments. By comparison, many of these methods achieve over 90\% R1 in urban VPR datasets, and AnyLoc achieves an average of 65\% R1 on unstructured environments such as subterranean, underwater, and in air-to-air VPR settings.

Fine-tuning provides inconsistent improvements with marginal gains, suggesting the primary bottleneck is not dataset-specific overfitting, but a fundamental lack of consistent visual correspondence between ground and aerial views. In particular, canopy occlusion removes most stable semantic landmarks, hindering appearance-based retrieval.

It is worth mentioning that re-ranking approaches for VPR may help reduce retrieval failures, but are inevitably limited by the top-N retrieval performance, and thus would provide minimal benefit without substantial improvements in Recall@N. Potential avenues to close the gap for VPR include utilising semantic information to infer tree trunk positions for local descriptor-based VPR (\eg Revisit Anything~\cite{gargRevisitAnythingVisual2025}), which may provide sufficiently robust geometry-informed cues to alleviate the severe domain gap between viewpoints. Additionally, leveraging more accessible HA airborne LiDAR as a 3D prior could assist the training of VPR models, which is discussed in \cref{sec:mmpr}.

\subsection{Cross-Modal Place Recognition}
\label{sec:cmpr}
Cross-modal place recognition poses some interesting questions.~For example, what are the best modalities for aligning ground and aerial viewpoints in forests? We use Lip-Loc~\cite{puligillaLIPLocLiDARImage2024} as our CMPR framework to evaluate this, though AGPlace~\cite{wangMultiModalAerialGroundCrossView2025} was also tested and failed to converge on \dataset{}.~In our experiments, we consider 3D products (\eg~LiDAR broadly available on UGVs) as the ground modality, and use RGB images as the primary aerial modality to reflect the ease-of-access to large-scale aerial imagery. Lip-Loc has been demonstrated to work reasonably in ground-to-ground forest PR using RGB images and LiDAR-derived range images cropped to match the image perspective~\cite{knightsWildCrossCrossModalLarge2026}. However, we argue this is ill-posed for ground-to-aerial PR, as the aerial perspective covers a \qty{360}{\degree} view of each ground query. We instead use full \qty{360}{\degree} range images to better align views. 

We also test other 3D modalities to evaluate how well they transfer for cross-modal alignment, including LiDAR-derived height maps, and swapping out Lip-Loc's image-based 3D encoder for a dedicated LiDAR encoder (HOTFormerLoc). We also use our DINOv3-based CHMv2~\cite{brandtCHMv2ImprovementsGlobal2026} canopy height maps as a source of aerial data to evaluate how well vision foundation models for depth/height estimation transfer to aerial orthorectified imagery in forests. See \cref{fig:hero} for comparisons of the modalities tested.

Cross-modal experiments (\cref{tab:cmpr}) confirm that 3D geometry is the most reliable cue in forested ground-to-aerial PR. Image-based 3D representations (range images, height maps) improve somewhat when paired with stronger vision backbones (DINOv2 and DINOv3), but still lag behind direct LiDAR representations produced by HOTFormerLoc. However, even the best configuration of Lip-Loc (DINOv2 + HOTFormerLoc LiDAR encoder) fails to reach 10\% R1 on average, placing cross-modal PR performance significantly below LPR methods. Compared to VPR approaches though, the addition of 3D cues leads to a gain over vision-only ground-to-aerial matching. These results suggest that while learned visual features can partially capture structural similarities (\eg, canopy density patterns or terrain contours), they remain insufficient to fully resolve the severe viewpoint-induced ambiguity in dense forests.

Surprisingly, range images consistently outperform height map-based approaches, even though the domain gap between aerial RGB and ground height maps is expected to be lower than \qty{360}{\degree} range images. We hypothesise this is due to fine-grained depth information being discarded when converting to height map format, and the limited vertical FoV of the LiDAR scanner preventing sampling of the canopy within several metres, reducing potential overlap.

Furthermore, CHMv2-predicted canopy height maps fail to significantly improve performance over aerial RGB, though they do perform consistently better on the unseen QCAT and Samford sequences. This indicates that 2D depth foundation models cannot yet replace true 3D geometry in forest environments, particularly for aerial orthorectified photos which are rarely present in pre-training data and lack the perspective cues expected by depth prediction networks.

\begin{table*}[t]
\centering
\renewcommand{\arraystretch}{0.95}
\resizebox{\linewidth}{!}{
\begin{tabular}{llcccccccccccccccccc}
\toprule

\multirow{2}{*}{Method (Image Encoder)} & \multicolumn{2}{c}{Modality} & \multicolumn{3}{c}{Karawatha} & \multicolumn{3}{c}{Venman} & \multicolumn{3}{c}{QCAT} & \multicolumn{3}{c}{Samford} & \multicolumn{3}{c}{Average} \\
& Ground & Aerial & R1 & R5 & R10 & R1 & R5 & R10 & R1 & R5 & R10 & R1 & R5 & R10 & R1 & R5 & R10  \\
\midrule
Lip-Loc~\cite{puligillaLIPLocLiDARImage2024} (ResNet50) & Range Img & RGB & 1.06 & 4.05 & \textbf{7.05} & \textbf{5.17} & \textbf{10.66} & \textbf{14.75} & 1.16 & 8.52 & 16.13 & 0.30 & 0.60 & 1.27 & 1.92 & 5.96 & 9.80 \\ 
Lip-Loc~\cite{puligillaLIPLocLiDARImage2024} (DINOv2) & Range Img & RGB & 1.67 & 3.91 & 6.36 & 1.16 & 3.28 & 5.35 & \textbf{14.45} & \textbf{36.26} & \textbf{50.19} & \textbf{0.90} & 1.95 & 3.22 & \textbf{4.55} & \textbf{11.35} & \textbf{16.28} \\ 
Lip-Loc~\cite{puligillaLIPLocLiDARImage2024} (DINOv3) & Range Img & RGB & \textbf{2.03} & \textbf{4.47} & 6.29 & 1.80 & 4.81 & 7.76 & 10.45 & 22.84 & 33.16 & 0.30 & \textbf{2.02} & \textbf{4.94} & 3.65 & 8.54 & 13.04 \\ 
\midrule
Lip-Loc~\cite{puligillaLIPLocLiDARImage2024} (ResNet50) & Height Map & RGB & 1.68 & \textbf{5.29} & \textbf{9.42} & \textbf{2.66} & \textbf{8.95} & \textbf{13.16} & 4.90 & 15.61 & 26.06 & \textbf{0.30} & \textbf{1.57} & 2.62 & 2.39 & 7.86 & 12.82 \\ 
Lip-Loc~\cite{puligillaLIPLocLiDARImage2024} (DINOv2) & Height Map & RGB & 0.66 & 2.31 & 3.82 & 0.99 & 3.74 & 5.77 & \textbf{14.06} & \textbf{27.35} & \textbf{39.10} & 0.07 & 0.97 & \textbf{3.97} & \textbf{3.95} & \textbf{8.59} & 13.17 \\ 
Lip-Loc~\cite{puligillaLIPLocLiDARImage2024} (DINOv3) & Height Map & RGB & \textbf{2.00} & 4.87 & 8.09 & 1.60 & 4.76 & 7.26 & 7.61 & 20.52 & 36.77 & 0.07 & 0.37 & 0.97 & 2.82 & 7.63 & \textbf{13.27} \\ 
\midrule
Lip-Loc~\cite{puligillaLIPLocLiDARImage2024} (ResNet50) & Height Map & CHMv2 & \textbf{0.50} & \textbf{1.76} & \textbf{3.46} & \textbf{0.24} & \textbf{0.73} & 1.22 & 12.65 & 41.16 & 55.48 & 0.30 & 1.65 & 3.60 & 3.42 & 11.33 & 15.94 \\ 
Lip-Loc~\cite{puligillaLIPLocLiDARImage2024} (DINOv2) & Height Map & CHMv2 & 0.18 & 1.24 & 2.51 & 0.11 & 0.70 & \textbf{1.36} & 12.65 & 31.74 & 43.61 & 0.30 & 2.92 & \textbf{4.87} & 3.31 & 9.15 & 13.09 \\ 
Lip-Loc~\cite{puligillaLIPLocLiDARImage2024} (DINOv3) & Height Map & CHMv2 & 0.44 & 1.65 & 2.85 & 0.19 & 0.60 & 1.18 & \textbf{18.45} & \textbf{45.03} & \textbf{59.10} & \textbf{1.05} & \textbf{3.15} & 4.04 & \textbf{5.03} & \textbf{12.61} & \textbf{16.79} \\ 
\midrule
Lip-Loc~\cite{puligillaLIPLocLiDARImage2024} (ResNet50)  & LiDAR & RGB & \textbf{2.29} & 5.23 & 6.96 & \textbf{4.12} & \textbf{10.15} & \textbf{14.12} & 10.19 & 27.21 & 41.96 & \textbf{2.05} & \textbf{7.71} & \textbf{11.57} & 4.66 & 12.58 & 18.65   \\
Lip-Loc~\cite{puligillaLIPLocLiDARImage2024} (DINOv2)  & LiDAR & RGB & 1.58 & 5.28 & \textbf{8.67} & 2.86 & 6.78 & 9.94 & \textbf{22.92} & \textbf{43.43} & \textbf{59.12} & 1.72 & 4.10 & 6.40 & \textbf{7.27} & \textbf{14.90} & \textbf{21.03}   \\
Lip-Loc~\cite{puligillaLIPLocLiDARImage2024} (DINOv3) & LiDAR & RGB & 2.15 & \textbf{5.51} & 8.01 & 1.94 & 6.80 & 11.23 & 13.78 & 35.24 & 48.63 & 0.53 & 1.96 & 3.40 & 4.60 & 12.38 & 17.82 \\

\bottomrule
\end{tabular}
}
\vspace{-.2 cm}
\caption{Cross-modal place recognition results on \dataset{}.}
\label{tab:cmpr}
\vspace{-5 mm}
\end{table*}

\begin{table}[t]
\centering
\renewcommand{\arraystretch}{1.15}
\setlength{\tabcolsep}{2.5pt}
\resizebox{\linewidth}{!}{
\normalsize
\begin{tabular}{lccccccccc}
\toprule
\multirow{2}{*}{Method} 
& \multicolumn{3}{c}{Karawatha} 
& \multicolumn{3}{c}{Venman} 
& \multicolumn{3}{c}{Average} \\
& R1 & R5 & R10 & R1 & R5 & R10 & R1 & R5 & R10 \\
\midrule
MinkLoc++~\cite{komorowskiMinkLocLidarMonocular2021} (RN18) 
& 29.4 & 48.3 & 59.3 & 29.0 & 51.1 & 62.0 & 29.2 & 49.7 & 60.7 \\
MinkLoc++v2 (RN50) 
& \textbf{48.2} & \textbf{69.7} & \textbf{77.4} & 37.7 & 60.7 & 71.0 & 43.0 & 65.2 & 74.2 \\
MinkLoc++v2 (DINOv2) 
& 46.7 & 68.6 & 77.1 & \textbf{48.9} & \textbf{76.8} & \textbf{86.8} & \textbf{47.8} & \textbf{72.7} & \textbf{81.9} \\
UMF~\cite{garcia-hernandezUnifyingLocalGlobal2024} (RN50) 
& 24.7 & 43.2 & 52.0 & 12.9 & 29.4 & 41.1 & 18.8 & 36.3 & 46.6 \\
\bottomrule
$^*$ RN means ResNet.
\end{tabular}
}
\vspace{-.3 cm}
\caption{Multi-modal place recognition results on \dataset{}.}
\label{tab:mmpr}
\vspace{-.2 cm}
\end{table}

\begin{table}[t]
\centering
\renewcommand{\arraystretch}{0.95}
\setlength{\tabcolsep}{2.2pt} %

\resizebox{\linewidth}{!}{
\normalsize
\begin{tabular}{llccccccccc}
\toprule
\multirow{2}{*}{Method} 
& \multirow{2}{*}{\makecell{GNSS\\Limit}} 
& \multicolumn{3}{c}{Image-Only} 
& \multicolumn{3}{c}{LiDAR-Only} 
& \multicolumn{3}{c}{Multi-Modal} \\
& & R1 & R5 & R10 & R1 & R5 & R10 & R1 & R5 & R10 \\
\midrule
MLoc++~\cite{komorowskiMinkLocLidarMonocular2021} (RN18) 
& -- 
& 1.0 & 2.4 & 3.4 
& 36.2 & 58.5 & 69.5 
& 29.2 & 49.7 & 60.7 \\

MLoc++v2 (RN50) 
& -- 
& 0.6 & 1.9 & 3.6 
& 50.4 & 72.1 & 80.0 
& 43.0 & 65.2 & 74.2 \\

MLoc++v2 (DINOv2) 
& -- 
& 1.0 & 4.0 & 7.0 
& \textbf{51.6} & \textbf{75.4} & \textbf{83.1} 
& \textbf{47.8} & \textbf{72.7} & \textbf{81.9} \\

MLoc++v2 (DINOv2)$^\ddagger$ 
& -- 
& \textbf{4.0} & \textbf{8.9} & \textbf{12.5} 
& 35.9 & 59.7 & 70.4 
& 41.7 & 65.8 & 76.0 \\
\midrule

MLoc++v2 (DINOv2) 
& 100\,m 
& 11.1 & 33.0 & 48.8 
& \textbf{73.6} & \textbf{91.6} & \textbf{95.7} 
& \textbf{70.9} & \textbf{90.0} & \textbf{95.1} \\

MLoc++v2 (DINOv2)$^\ddagger$ 
& 100\,m 
& \textbf{16.5} & \textbf{38.8} & \textbf{53.3} 
& 65.9 & 85.5 & 91.9 
& 67.0 & 86.4 & 92.4 \\
\midrule

MLoc++v2 (DINOv2) 
& 50\,m 
& 28.2 & 73.1 & 90.7 
& \textbf{84.5} & \textbf{98.3} & \textbf{99.9} 
& \textbf{81.8} & \textbf{98.2} & \textbf{99.8} \\

MLoc++v2 (DINOv2)$^\ddagger$ 
& 50\,m 
& \textbf{32.9} & \textbf{76.5} & \textbf{92.5} 
& 79.6 & 96.9 & 99.8 
& 79.8 & 97.3 & \textbf{99.8} \\
\bottomrule
\multicolumn{11}{l}{$^\ddagger$ indicates modality dropout augmentation was used.}
\end{tabular}
}
\vspace{-.3 cm}
\caption{Performance analysis of each modality branch in MinkLoc++ (MLoc++), with pseudo-GNSS constraints on the query search space. Average results over all sequences are reported.}
\label{tab:modality_comparison}
\vspace{-6 mm}
\end{table}

\vspace{-1 mm}
\subsection{Multi-Modal Place Recognition}
\label{sec:mmpr}

Multi-modal learning promises to leverage the complementary relationships between disparate modalities to improve performance beyond what uni-modal representations can achieve.~To evaluate MMPR under the severe domain gaps seen in \dataset{}, we evaluate two SOTA approaches with open-source implementations: MinkLoc++~\cite{komorowskiMinkLocLidarMonocular2021} for late fusion and UMF~\cite{garcia-hernandezUnifyingLocalGlobal2024} for mid fusion.~AdaFusion~\cite{laiAdaFusionVisualLiDARFusion2022} was also tested, but failed to converge.~As MinkLoc++ originally used lightweight ResNet18 and MinkLoc3D encoders, we also test a modernised version equipped with ResNet50\,/\,DINOv2~\cite{oquabDINOv2LearningRobust2023} image encoders and a MinkLoc3Dv2~\cite{komorowskiImprovingPointCloud2022} LiDAR encoder, denoted MinkLoc++v2.

Multi-modal results (\cref{tab:mmpr}) show combining LiDAR and image modalities provides consistent gains over image-only baselines.~MinkLoc++v2 with DINOv2 image backbone achieves the strongest multi-modal performance, reaching 47.8\%\,/\,72.7\%\,/\,81.9\% average R1\,/\,R5\,/\,R10.
However, LiDAR-only methods such as HOTFLoc++ remain stronger overall, indicating current fusion strategies do not fully exploit the complementary nature of modalities in this highly occluded setting.

To investigate the contribution of each modality in MMPR, we evaluate the uni-modal retrieval performance of the image and LiDAR branches of MinkLoc++ in \cref{tab:modality_comparison}. As expected from VPR and CMPR results, the image-only branch contributes little to retrieval performance, achieving only 1.0\% R1 with the DINOv2 variant of MinkLoc++v2. Interestingly, the poor image branch performance seems to harm multi-modal performance, with the LiDAR-only branch of the same model achieving 51.6\% R1: 3.8 percentage points better than the full multi-modal descriptor.

To try improve the robustness of MinkLoc++, we apply modality dropout~\cite{neverovaModDropAdaptiveMultiModal2016} during training to force the model to learn multi-modal representations that remain distinct when one modality is missing or degraded. This improves the image-only R1\,/\,R5\,/\,R10 to 4.0\%\,/\,8.9\%\,/\,12.5\%, but has the undesired effect of reducing both LiDAR-only and multi-modal absolute performance. However, the multi-modal descriptor now performs better than LiDAR-only, indicating reduced dependence on the LiDAR modality.

Notably though, the process of multi-modal training extracts better representations from RGB images than any of the VPR approaches.~Comparing MinkLoc++v2 (DINOv2)$^\ddagger$ with the overall best VPR approach, AnyLoc, we see that MinkLoc++v2 achieves a nearly $3\times$ improvement in Recall@N.
This indicates that 3D geometry could be used to ground the training of VPR models, by adding geometric constraints unobtainable from RGB alone. Airborne LiDAR could be a powerful source of data for such grounding, as its relative affordability and scalability could enable large-scale pre-training of 3D-aware VPR models.

\begin{figure*}[t]
    \centering
    \setlength{\fboxrule}{1.2pt}
    \setlength{\fboxsep}{0pt}
    \begin{subfigure}[b]{0.22\linewidth}
        \centering
        \color{black}\fbox{\includegraphics[width=\linewidth]{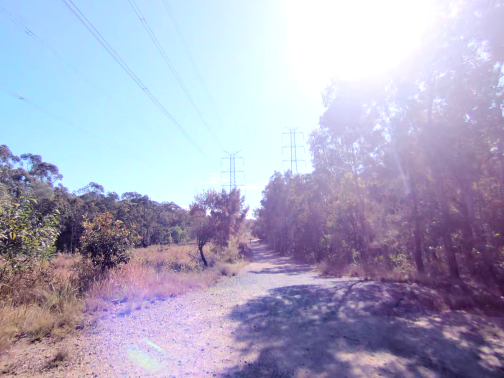}}
        \color{black}\fbox{\includegraphics[width=\linewidth]{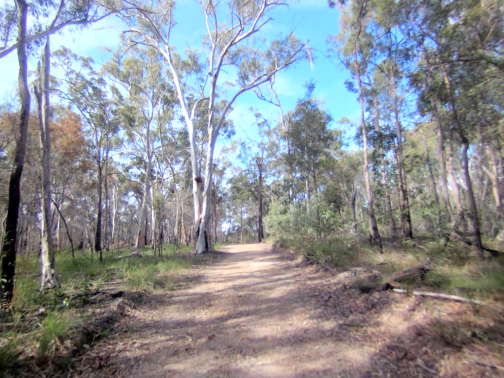}}
        \caption{Image Query}
        \label{fig:im_lidar_im_q}
    \end{subfigure}
    \hfill
    \begin{subfigure}[b]{0.22\linewidth}
        \centering
        \color{green}\fbox{\includegraphics[width=\linewidth]{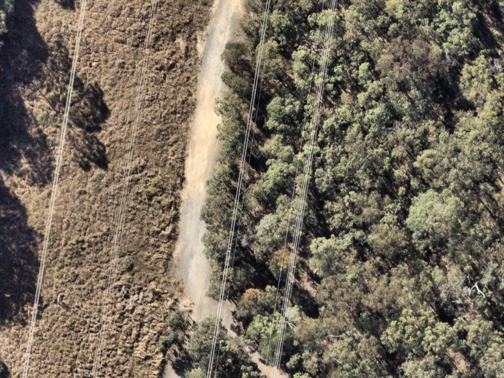}}
        \color{green}\fbox{\includegraphics[width=\linewidth]{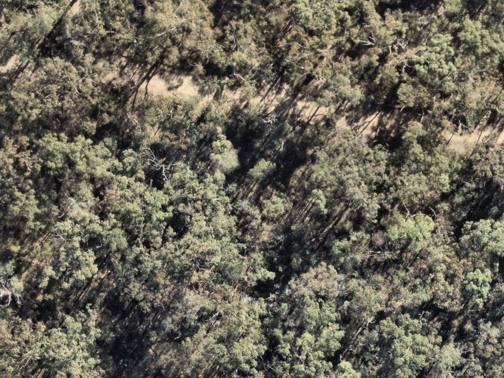}}
        \caption{Top-1 Candidate ($<\,$\qty{30}{\m})}
        \label{fig:im_lidar_im_tp}
    \end{subfigure}
    \hfill
    \begin{subfigure}[b]{0.22\linewidth}
        \centering
        \color{black}\fbox{\includegraphics[width=\linewidth]{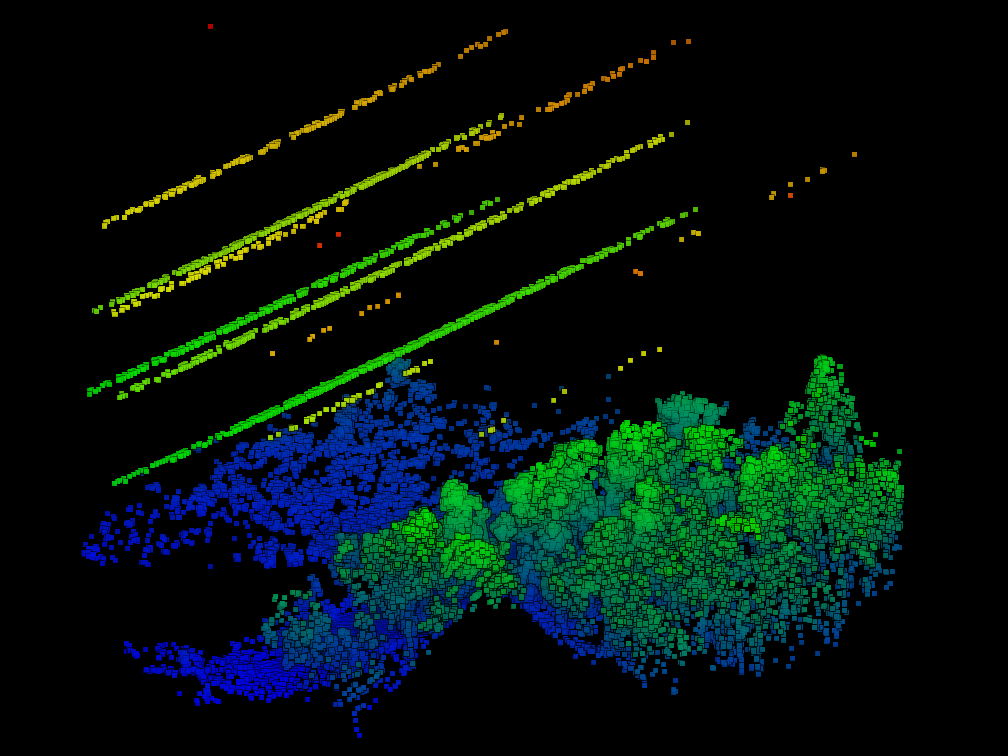}}
        \color{black}\fbox{\includegraphics[width=\linewidth]{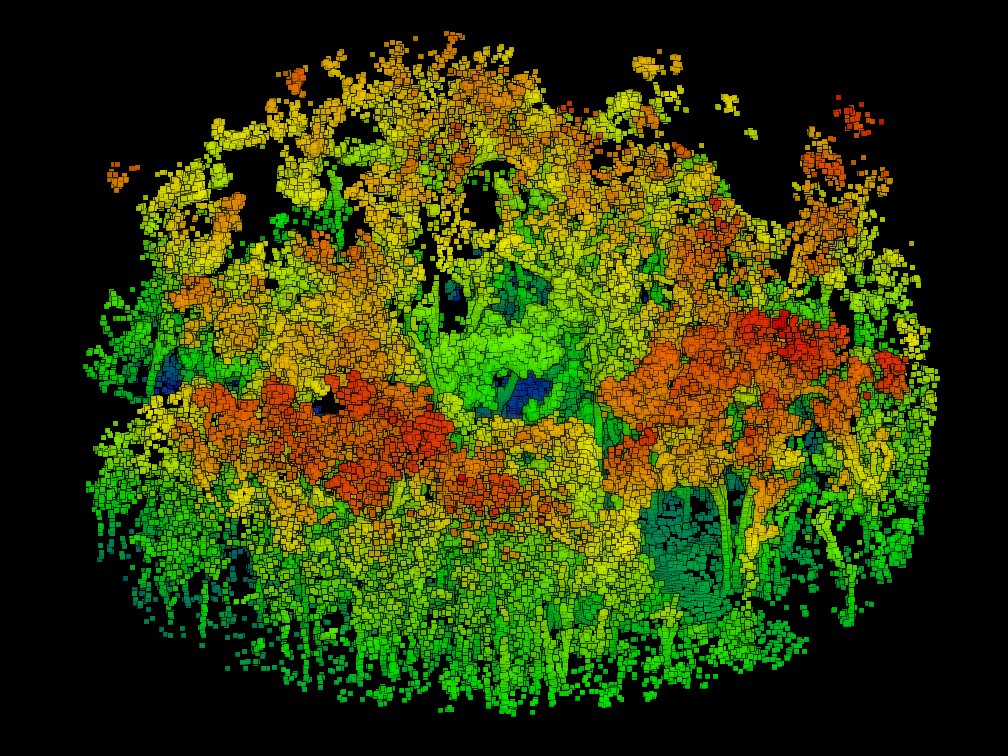}}
        \caption{LiDAR Query}
        \label{fig:im_lidar_lidar_q}
    \end{subfigure}
    \hfill
    \begin{subfigure}[b]{0.22\linewidth}
        \centering
        \color{red}\fbox{\includegraphics[width=\linewidth]{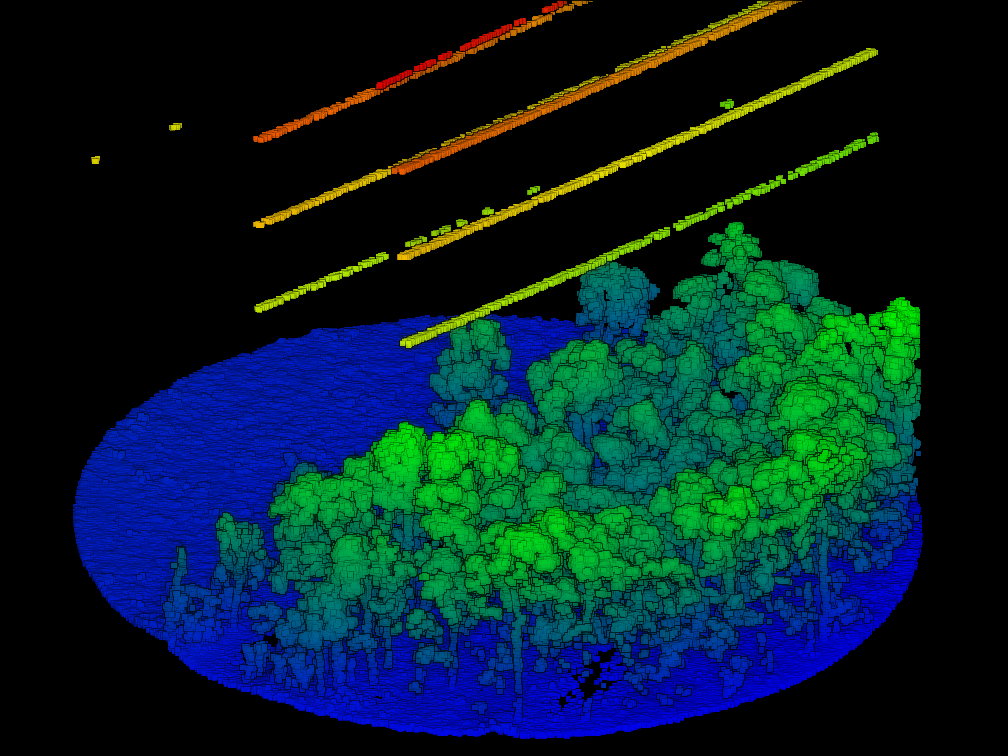}}
        \color{red}\fbox{\includegraphics[width=\linewidth]{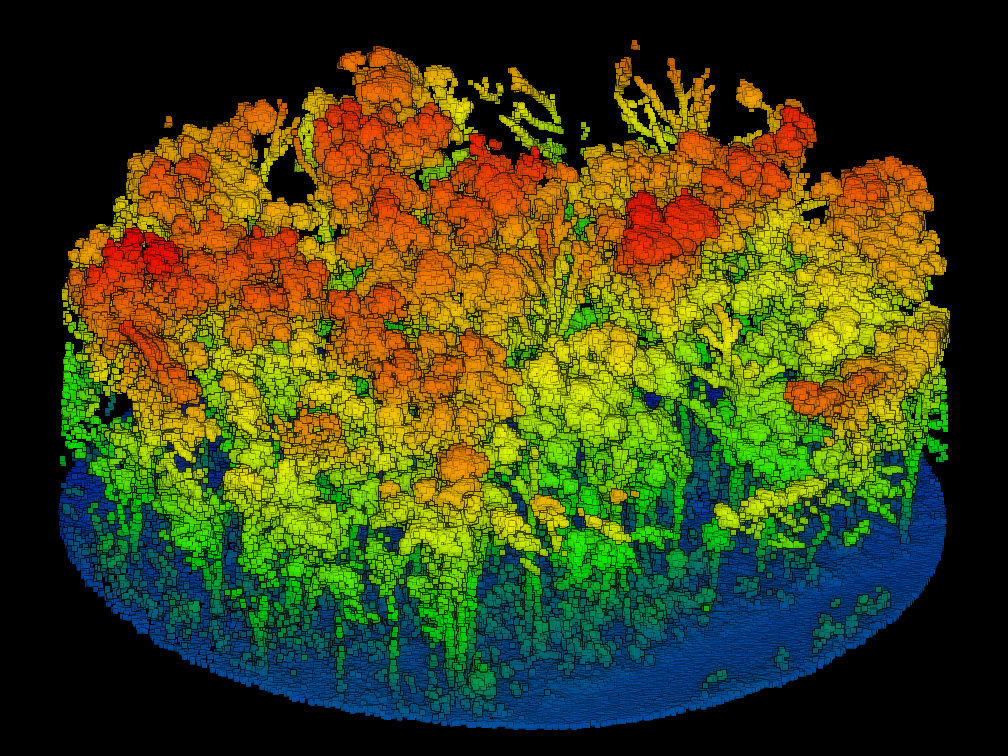}}
        \caption{{Top-1 Candidate ($>\,$\qty{200}{\m})}}
        \label{fig:im_lidar_lidar_fp}
    \end{subfigure}
    \vspace{-2 mm} 
    \caption{Queries where MinkLoc++v2 (DINOv2) image-only branch successfully localised where LiDAR-only did not. RGB captures ground texture details that are invisible to LiDAR, demonstrating that RGB can provide useful information in forests if integrated correctly.
    }
    \label{fig:im_lidar_success}
\vspace{-5 mm}    
\end{figure*}

\subsection{Place Recognition with coarse GNSS priors}
\label{sec:gps_prior}
The pseudo-GNSS evaluation (\cref{tab:modality_comparison}) provides insights into how spatial priors affect PR under extreme occlusion and viewpoint change. Here, we assume that GNSS is available under the canopy for coarse positioning, and simulate 3-$\sigma$ error thresholds of \qty{100}{\m} and \qty{50}{\m} for verifying retrievals. As search space is reduced by this constraint, performance improves dramatically across all modalities. For example, MinkLoc++v2 (DINOv2) improves from 51.6\% R1 unconstrained LiDAR-only retrieval to 84.5\% R1 under a \qty{50}{\m} constraint, with similar gains observed in the multi-modal branch. The near-100\% R10 implies that a suitable re-ranking approach could achieve flawless performance. This suggests that many failures in unconstrained settings arise from global perceptual aliasing rather than local ambiguity.

However, performance gains are more modest for the image-only branch of MinkLoc++v2, improving from 4.0\% R1 to only 32.9\% R1 under the \qty{50}{\m} constraint. Whilst this is a large relative improvement in performance, the low absolute improvement highlights that vision-only approaches cannot reliably identify viewpoint-agnostic correspondences even when perceptual aliasing is eliminated.

\subsection{Discussion}
\label{sec:discussion}

\dataset{}'s setting differs substantially from urban cross-view localisation, where road layouts and building footprints can provide strong geometric and semantic cues. In forests, repeated vegetation patterns, canopy occlusions, seasonal appearance changes, weak structural landmarks, and limited visibility make both visual and LiDAR matching significantly more challenging. Moreover, approaches relying on ground spherical imagery or BEV projections to align viewpoints~\cite{ferversUncertaintyAwareVisionBasedMetric2023} are unusable in dense forests, where the under-canopy perspective is heavily occluded and lacks the stable urban layout assumptions typically exploited by these methods. As a result of the above, results show a significant gap between image- and LiDAR-based PR performance.

So, does RGB imagery carry no usable information compared to LiDAR for place recognition in forest environments? \Cref{fig:im_lidar_success} would suggest otherwise, as there are several cases where the image-only branch of MinkLoc++ successfully localises where the LiDAR-only branch fails. In particular, when the ground can be partially viewed from above, ground texture and colour provide a disambiguating signal that is unobtainable from LiDAR in which different ground surfaces are nearly indistinguishable. By contrast, the LiDAR-only branch is distracted by candidates which share geometric similarities, but do not correspond with the query location.
This demonstrates that visual features provide complementary information to LiDAR, particularly where geometric structure alone is insufficient. 

However, it is clear from \cref{tab:mmpr,tab:modality_comparison} that current fusion approaches are not adaptive enough to exploit RGB when it carries useful signal whilst ignoring it when it acts as a distractor. Even mid-fusion approaches like UMF struggle to achieve close to the performance of the late-fusion MinkLoc++, highlighting that cross-attention-based fusion is too reliant on strong modal alignment to exploit complementary information.~Robust approaches that can handle severe modal misalignment remain an interesting future research direction, and will likely be essential in creating reliable localisation systems for highly occluded environments.

One interesting direction for future research is the application of foundation monocular depth estimation models to estimate 3D structure from aerial RGB imagery. Initial experiments in \cref{sec:cmpr} indicated that current feed-forward models struggle to produce reliable or high fidelity depth estimations from aerial imagery. A potential avenue for future work is to leverage the HA LiDAR data of \dataset{} to fine-tune such models for aerial perspective views. As discussed previously, HA LiDAR offers a relatively inexpensive way to obtain large-scale aerial 3D data, which can enable foundation model pre-training.

%% file: sec/5_conclusion.tex
\section{Conclusion}
\label{sec:conclusion}

This paper introduces \dataset{}, a large-scale multi-modal, multi-session benchmark designed to advance ground-to-aerial PR and localisation in forests.~With aligned ground--aerial RGB and LiDAR, \dataset{} enables systematic evaluation of cross-view, cross- and multi-modal localisation under realistic conditions. Extensive baselines highlight the fundamental difficulty of ground-to-aerial PR in forests. LiDAR-based methods are most robust to viewpoint changes and occlusion, while vision-only approaches struggle due to limited appearance overlap and perceptual aliasing. Although multi-modal approaches offer some improvements, current fusion strategies remain insufficient to fully exploit complementary information.
The findings underscore the need for better alignment between RGB and LiDAR representations, adaptive fusion, and scalable aerial LiDAR to improve image-derived geometry suitable for PR.
We hope \dataset{} will serve as a useful benchmark for advancing robust localisation and autonomy in forests.

\section*{Acknowledgements}
This work was conducted as part of a joint CSIRO-QUT PhD program. The authors acknowledge the support of Nearmap in providing high-resolution geo-referenced aerial imagery, and the Queensland Government in providing the airborne LiDAR scans. They also appreciate continuous support from CSIRO Robotics.

%% file: sec/X_supplementary.tex
\clearpage
\setcounter{figure}{0}
\renewcommand{\thefigure}{S\arabic{figure}}
\setcounter{table}{0}
\renewcommand{\thetable}{S\arabic{table}}
\setcounter{equation}{0}
\renewcommand{\theequation}{S\arabic{equation}}
\setcounter{section}{0}
\renewcommand{\thesection}{S\arabic{section}}
\maketitlesupplementary

\section{Model-Specific Hyperparameters}
We provide a summary of the main model-specific hyperparameters used to train methods on \dataset{} in \cref{tab:sup_hyperparams}. Any hyperparameters not mentioned can be assumed as set to the defaults provided for each model.

\begin{table}[t]
    \centering
    \renewcommand{\arraystretch}{1.15}
    \resizebox{\linewidth}{!}{
    \begin{tabular}{llccccl}
    \toprule
    & Model & LR & Epoch & BS & Res (W, H) & Other\\
    \midrule
    \multirow{6}{*}{\rotatebox{90}{LPR}} & MinkLoc3Dv2~\cite{komorowskiImprovingPointCloud2022} & $1\mathrm{e}^{-3}$ & 120 & 2048 & -- & \\
    & EgoNN~\cite{komorowskiEgoNNEgocentricNeural2022} & $1\mathrm{e}^{-3}$ & 120 & 128 & -- & \\
    & LoGG3D-Net~\cite{vidanapathiranaLoGG3DNetLocallyGuided2022} & $1\mathrm{e}^{-3}$ & 50 & 12 & -- & Feat dim: 32 \\
    & CrossLoc3D~\cite{guanCrossLoc3DAerialGroundCrossSource2023} & $5\mathrm{e}^{-3}$ & 200 & 128 & -- & \\
    & HOTFormerLoc~\cite{griffithsHOTFormerLocHierarchicalOctree2025} & $8\mathrm{e}^{-4}$ & 100 & 2048 & -- & \\
    & HOTFLoc++~\cite{griffithsHOTFLocEndtoEndHierarchical2025} & $8\mathrm{e}^{-4}$ & 60 & 256 & -- & \\
    \midrule
    \multirow{5}{*}{\rotatebox{90}{VPR}} & NetVLAD~\cite{arandjelovicNetVLADCNNArchitecture2016} & $1\mathrm{e}^{-4}$ & 40 & 64 & (320, 320) & \\
    & MixVPR~\cite{ali-beyMixVPRFeatureMixing2023} & $1\mathrm{e}^{-4}$ & 40 & 128 & (320, 320) & \\
    & SALAD~\cite{izquierdoOptimalTransportAggregation2024} & $1\mathrm{e}^{-4}$ & 40 & 128 & (322, 322) & \\
    & BoQ~\cite{ali-beyBoQPlaceWorth2024} & $1\mathrm{e}^{-4}$ & 40 & 128 & (322, 322) & \\
    & AnyLoc~\cite{keethaAnyLocUniversalVisual2024} & -- & -- & -- & (336, 336) & Domain: global \\
    \midrule
    \multirow{2}{*}{\rotatebox{90}{CMPR}} & Lip-Loc~\cite{puligillaLIPLocLiDARImage2024} (RN50) & $1\mathrm{e}^{-4}$ & 50 & 128 & (288, 288) & Range Img: (896, 224)\\
    & Lip-Loc~\cite{puligillaLIPLocLiDARImage2024} (DINOv\underline{\ \ }) & $1\mathrm{e}^{-4}$ & 50 & 128 & (336, 336) & Range Img: (896, 224)\\
    \midrule
    \multirow{4}{*}{\rotatebox{90}{MMPR}} & MinkLoc++~\cite{komorowskiMinkLocLidarMonocular2021} (RN18) & $1\mathrm{e}^{-3}$ & 100 & 160 & (224, 224) & Image LR: $1\mathrm{e}^{-4}$ \\
    & MinkLoc++v2 (RN50) & $1\mathrm{e}^{-3}$ & 100 & 160 & (288, 288) & Image LR: $1\mathrm{e}^{-4}$ \\
    & MinkLoc++v2 (DINOv2) & $1\mathrm{e}^{-3}$ & 100 & 160 & (336, 336) & Image LR: $1\mathrm{e}^{-4}$ \\
    & UMF~\cite{garcia-hernandezUnifyingLocalGlobal2024} & $1\mathrm{e}^{-4}$ & 200 & 160 & (288, 288) & Only global feat  \\
    \bottomrule
    \multicolumn{7}{l}{LR = learning rate (Python notation), BS = batch size, Res = img resolution}\\
    \end{tabular}
    }
    \vspace{-2mm}
    \caption{Model-specific hyperparameters used on \dataset{}.}
    \label{tab:sup_hyperparams}
    \vspace{-2mm}
\end{table}

\section{Additional \dataset{} Information}
In this section, we provide additional details and information about the \dataset{} dataset, including visualisations and details of each forest, comparisons of our high altitude (HA) and low altitude (LA) LiDAR submaps, and additional information about our training and testing splits.

\subsection{\dataset{} Sequences}
\begin{figure*}[t]
    \centering
    \begin{subfigure}[b]{0.49\linewidth}
        \centering
        \includegraphics[width=\linewidth]{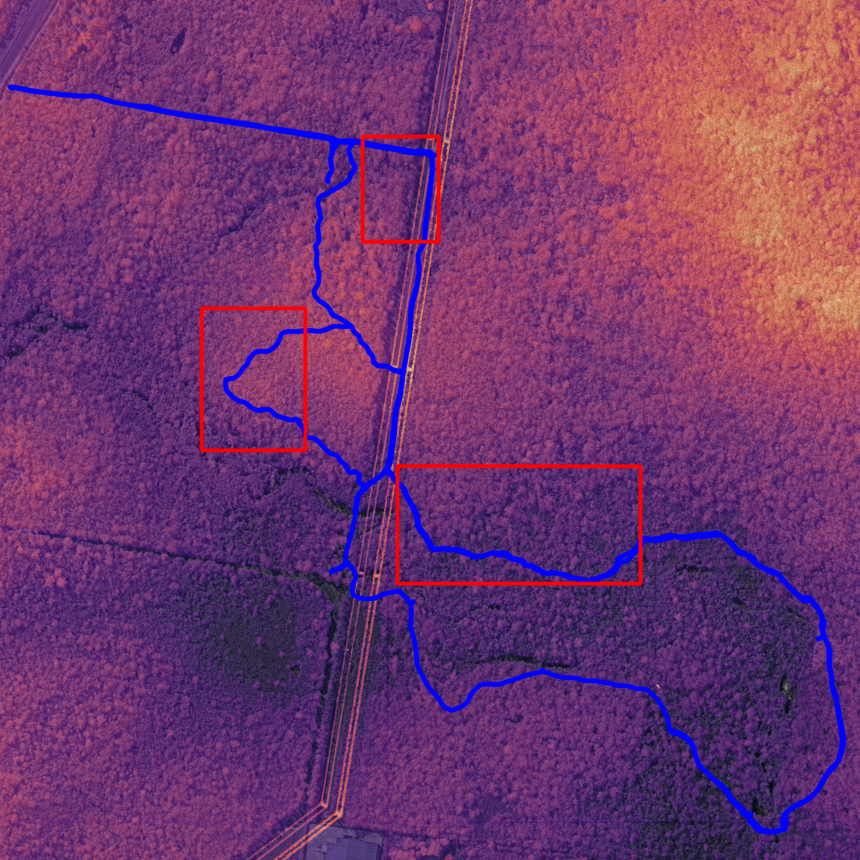}
        \caption{Karawatha}
    \end{subfigure}
    \hfill
    \begin{subfigure}[b]{0.49\linewidth}
        \centering
        \includegraphics[width=\linewidth]{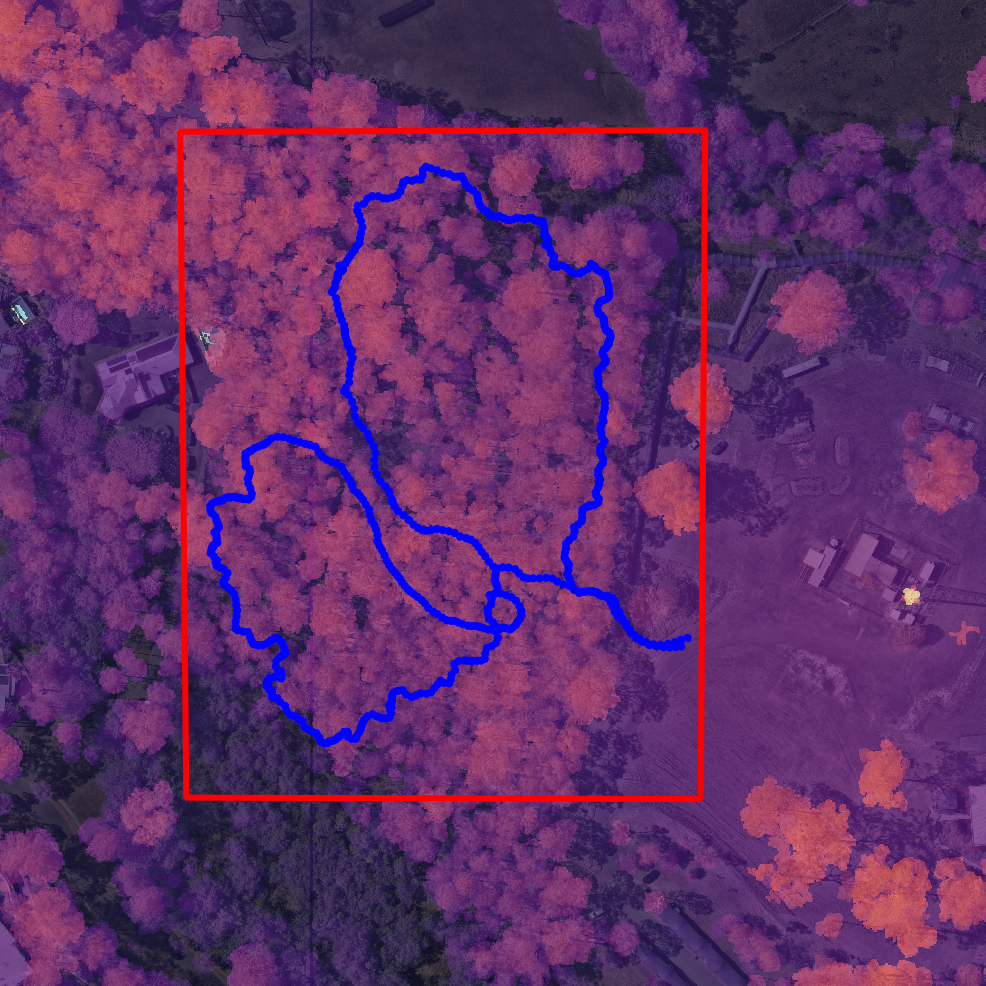}
        \caption{QCAT}
    \end{subfigure}
    \vfill
    \begin{subfigure}[b]{0.49\linewidth}
        \centering
        \includegraphics[width=\linewidth]{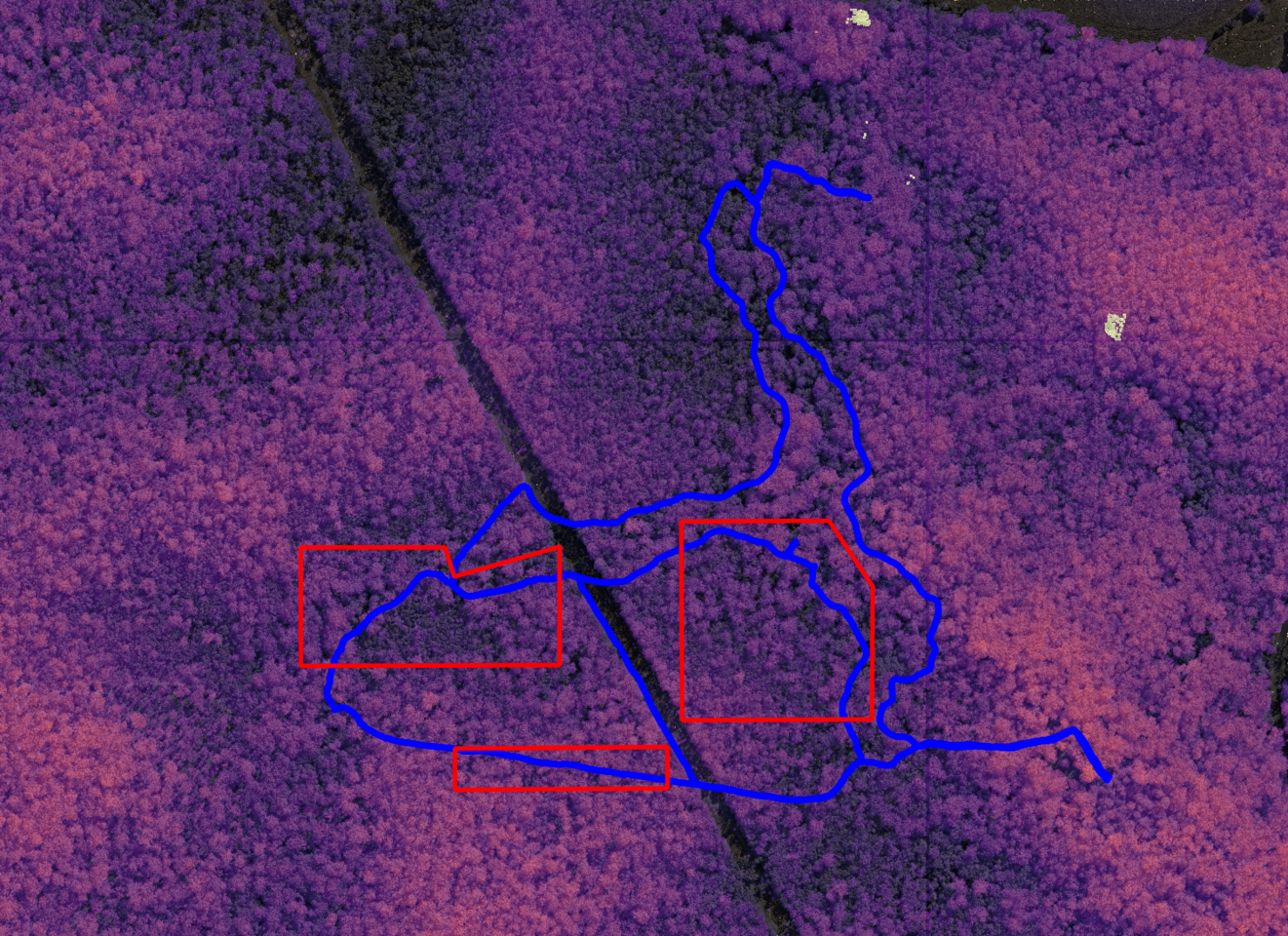}
        \caption{Venman}
    \end{subfigure}
    \hfill
    \begin{subfigure}[b]{0.49\linewidth}
        \centering
        \includegraphics[width=\linewidth]{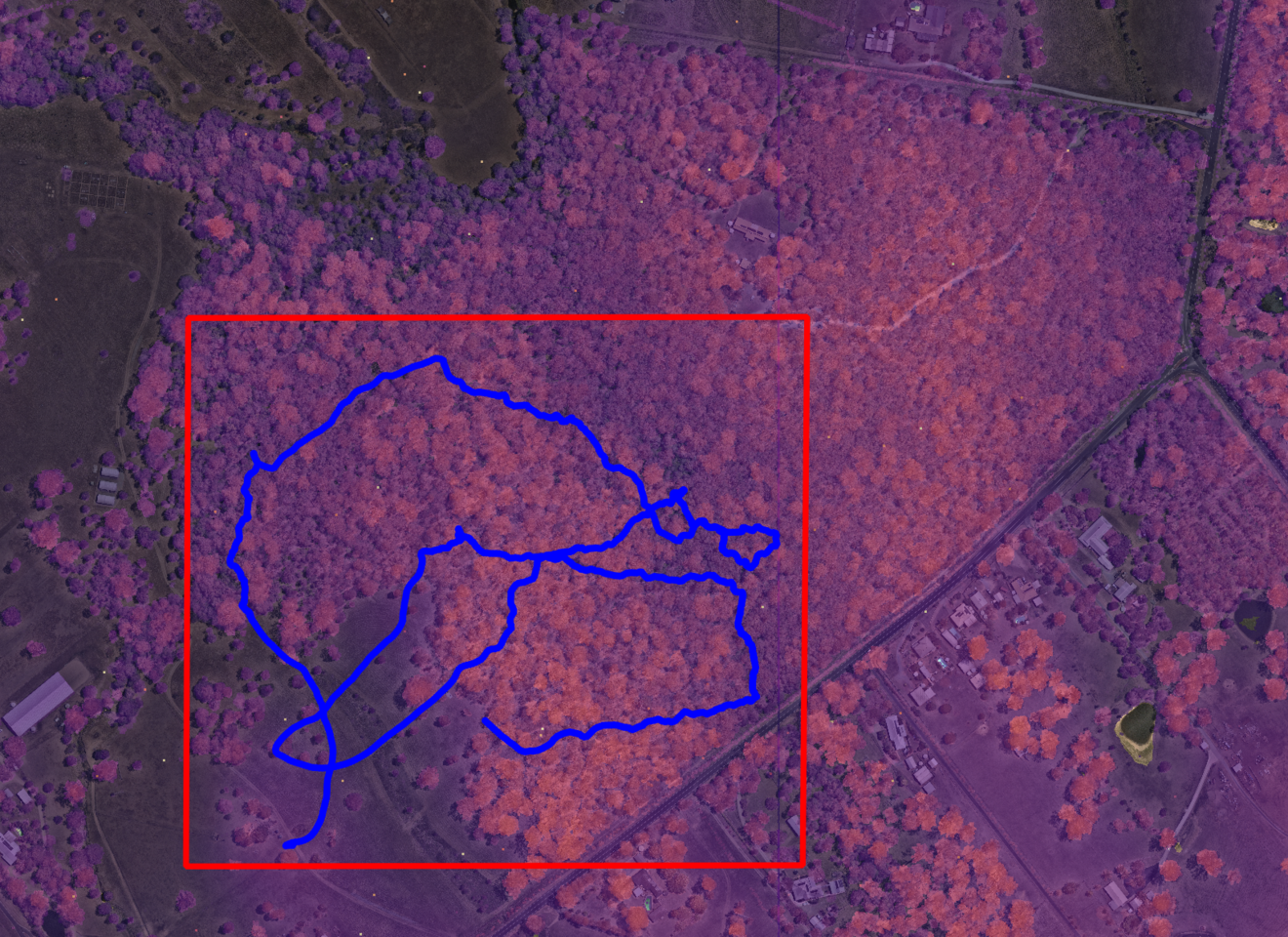}
        \caption{Samford}
    \end{subfigure}
    \caption{Overview of each forest in \dataset{}. Aerial RGB imagery is overlaid with ground trajectory (blue), test regions (red), and high altitude LiDAR scans colourised by z coordinate.}
    \label{fig:sup_test_region}
\end{figure*}

We provide an overview of all \dataset{} sequences in \cref{tab:sup_sequences}. Our unified benchmark introduces \qty{70.3}{\kilo\nothing} high-resolution aerial images and \qty{35.1}{\kilo\nothing} LiDAR submaps captured from high altitude, enhancing the ground RGB images, LiDAR submaps, and low altitude aerial submaps of WildCross~\cite{knightsWildCrossCrossModalLarge2026} and CS-Wild-Places~\cite{griffithsHOTFormerLocHierarchicalOctree2025}. A visualisation of all four forests and testing regions is provided in \cref{fig:sup_test_region}.

\begin{table}[t]
    \centering
    \renewcommand{\arraystretch}{1.15}
    \resizebox{\linewidth}{!}{
    \begin{tabular}{lcccccc}
    \toprule
    Forest & Sequence & Coverage & \# Images & \# Submaps \\
    \midrule
    \multirow{6}*{Karawatha} & \texttt{2019\_06\_11\_airborne}$^*$ & 1.80\,km$^2$ & 17792$^*$ & 17792$^*$ \\
    & \texttt{2021\_06\_21\_K-02\_ground} & 5.66\,km & 75570 & 10075 \\
    & \texttt{2021\_06\_22\_K-01\_ground}  & 5.14\,km & 66120 & 8816 \\
    & \texttt{2021\_12\_13\_K-03\_ground}  & 6.27\,km & 14205 & 15150 \\
    & \texttt{2022\_08\_11\_K-04\_ground}  & 2.81\,km & 43766 & 5805 \\
    & \texttt{2024\_07\_10\_aerial}$^*$ & 1.80\,km$^2$ & 17792$^*$ & 17792 \\
    \midrule
    \multirow{6}*{Venman} & \texttt{2021\_06\_11\_V-01\_ground} & 2.64\,km & 35305 & 4699 \\
    & \texttt{2021\_06\_11\_V-02\_ground} & 2.64\,km & 34163 & 4557 \\
    & \texttt{2021\_12\_16\_V-03\_ground} & 4.59\,km & 63751 & 8470 \\
    & \texttt{2022\_06\_12\_airborne}$^*$ & 1.30\,km$^2$ & 12384$^*$ & 12384$^*$ \\
    & \texttt{2022\_08\_12\_V-04\_ground} & 2.81\,km & 43124 & 5739 \\
    & \texttt{2024\_07\_09\_aerial}$^*$ & 1.30\,km$^2$ & 12384$^*$ & 12384 \\
    \midrule
    \multirow{3}*{QCAT} & \texttt{2019\_06\_11\_airborne}$^*$ & 0.06\,km$^2$ & 382$^*$ & 382$^*$ \\
    & \texttt{2023\_09\_07\_ground} & 0.90\,km & -- & 775 \\
    & \texttt{2023\_09\_08\_aerial}$^*$ & 0.06\,km$^2$ & 382$^*$ & 382 \\
    \midrule
    \multirow{3}*{Samford} & \texttt{2018\_11\_25\_airborne}$^*$ & 0.58\,km$^2$ & 4568$^*$ & 4568$^*$ \\
    & \texttt{2023\_09\_17\_ground} & 2.60\,km & -- & 1335 \\
    & \texttt{2023\_09\_18\_aerial}$^*$ & 0.58\,km$^2$ & 4568$^*$ & 4568 \\
    \midrule
    \multirow{3}*{Total} & Ground & 36.06\,km & 376004 & 65422 \\
    & Aerial (LA) & 3.74\,km$^2$ & 35126 & 35126 \\
    & Aerial (HA) & 3.74\,km$^2$ & 35126 & 35126 \\

    \bottomrule
    \multicolumn{5}{l}{$^*$ indicates novel data layers introduced by our dataset.}\\
    \end{tabular}
    }
    \vspace{-2mm}
    \caption{Summary of \dataset{} sequences. 
    }
    \label{tab:sup_sequences}
    \vspace{-2mm}
\end{table}

\subsection{CHMv2 Height Map Comparison}
\begin{figure*}
    \centering
    \resizebox{0.75\linewidth}{!}{
    \begin{tabular}{c}
        \includegraphics[width=0.99\linewidth,trim={10 14 10 0},clip]{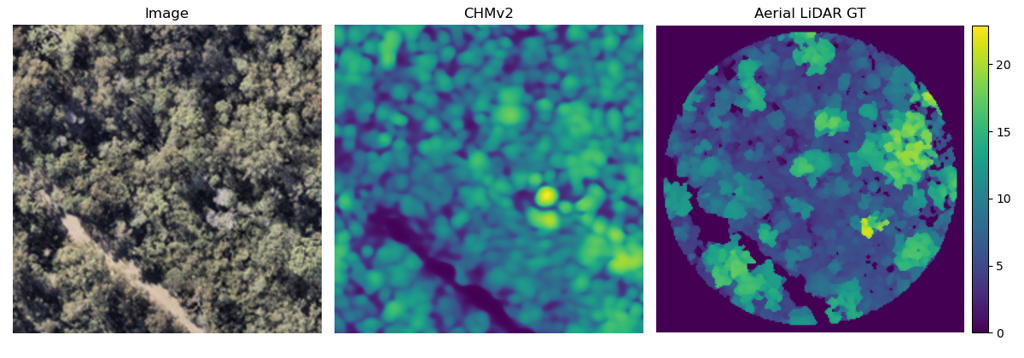} \\[2pt]
        \includegraphics[width=0.99\linewidth,trim={10 18 10 22},clip]{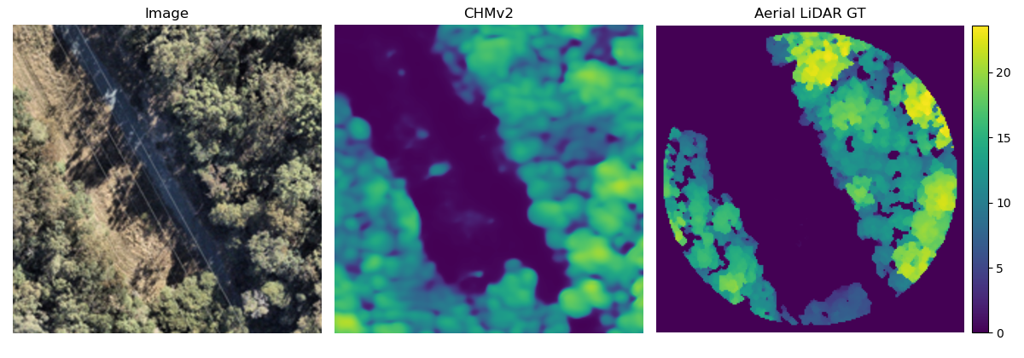} \\[2pt]
        \includegraphics[width=0.99\linewidth,trim={10 18 10 22},clip]{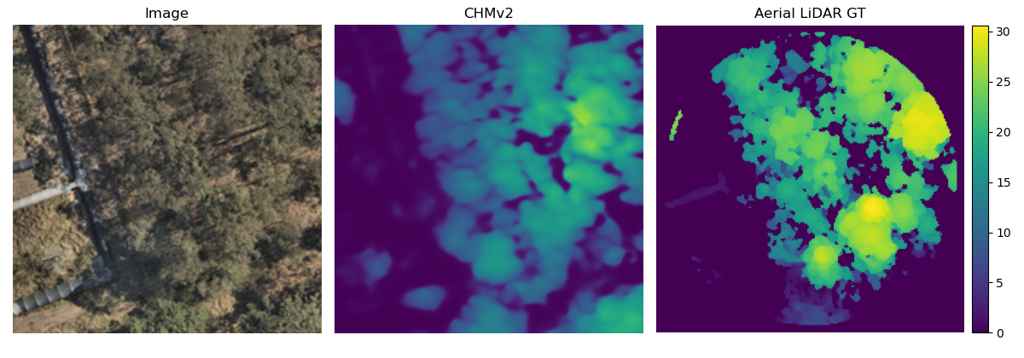} \\[2pt]
        \includegraphics[width=0.99\linewidth,trim={10 18 10 22},clip]{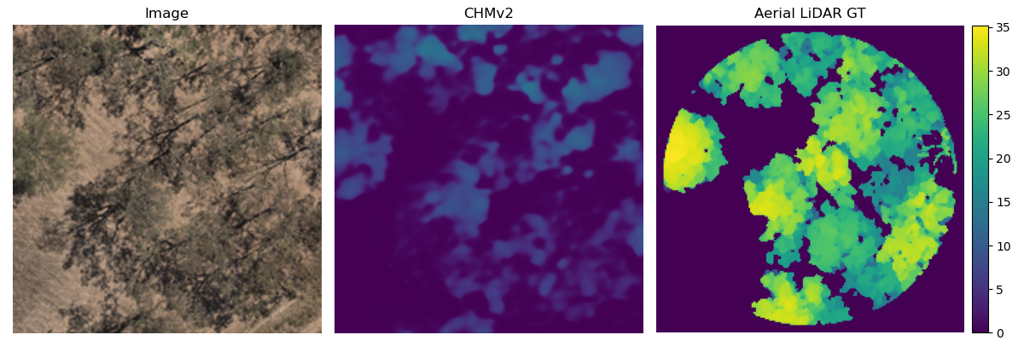} \\[2pt]
    \end{tabular}
    }
    \vspace{-2mm}
    \caption{Comparison of CHMv2~\cite{brandtCHMv2ImprovementsGlobal2026} height maps \vs LA aerial LiDAR ground truth. From top to bottom: Karawatha, Venman, QCAT, Samford.}
    \label{fig:sup_chmv2_comparison}
    \vspace{-2mm}
\end{figure*}

We provide comparisons between CHMv2~\cite{brandtCHMv2ImprovementsGlobal2026} predicted height maps and ground truth height maps derived from LA aerial LiDAR submaps in \cref{fig:sup_chmv2_comparison}. The CHMv2 height maps generally capture the rough contour and shape of the forest canopy, but struggle to accurately capture the fine-grained height variations captured by aerial LiDAR. Notably, CHMv2 fails catastrophically on the visually deceptive sparse trees seen in Samford, highlighting the bias of the model towards dense forested regions.

\subsection{Comparison of Low and High Altitude LiDAR}

\begin{figure*}
    \resizebox{\linewidth}{!}{
    \begin{tabular}{cccc}
        \renewcommand{\arraystretch}{1}
        \includegraphics[width=0.23\linewidth]{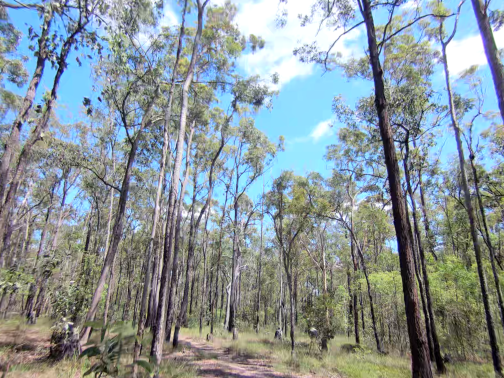} & \includegraphics[width=0.23\linewidth]{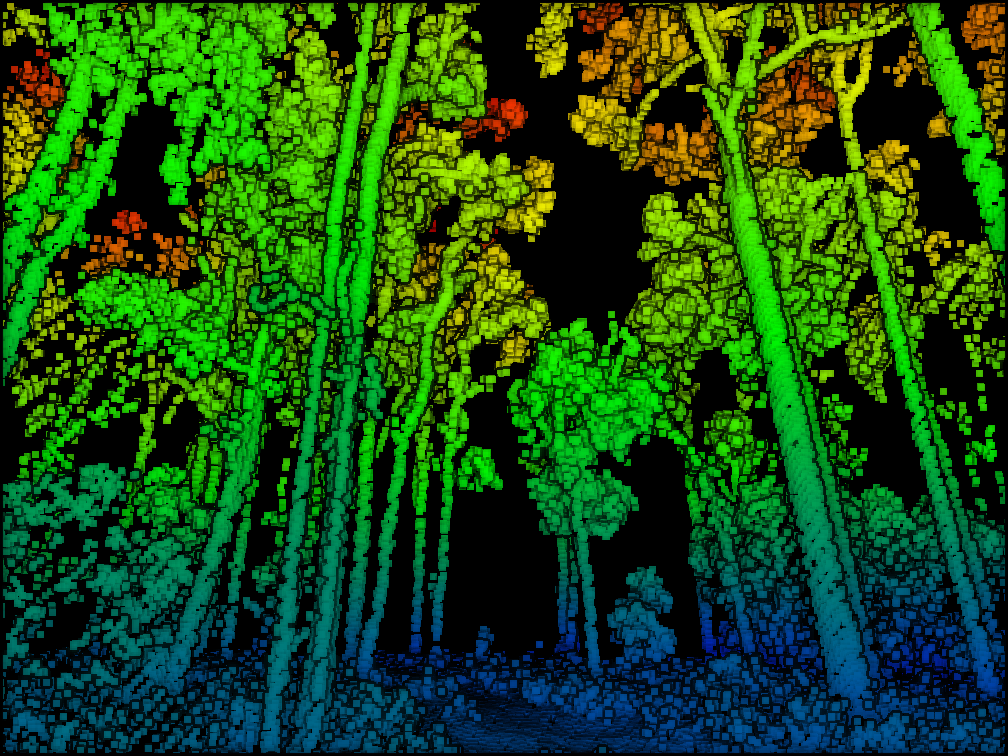} & \includegraphics[width=0.23\linewidth]{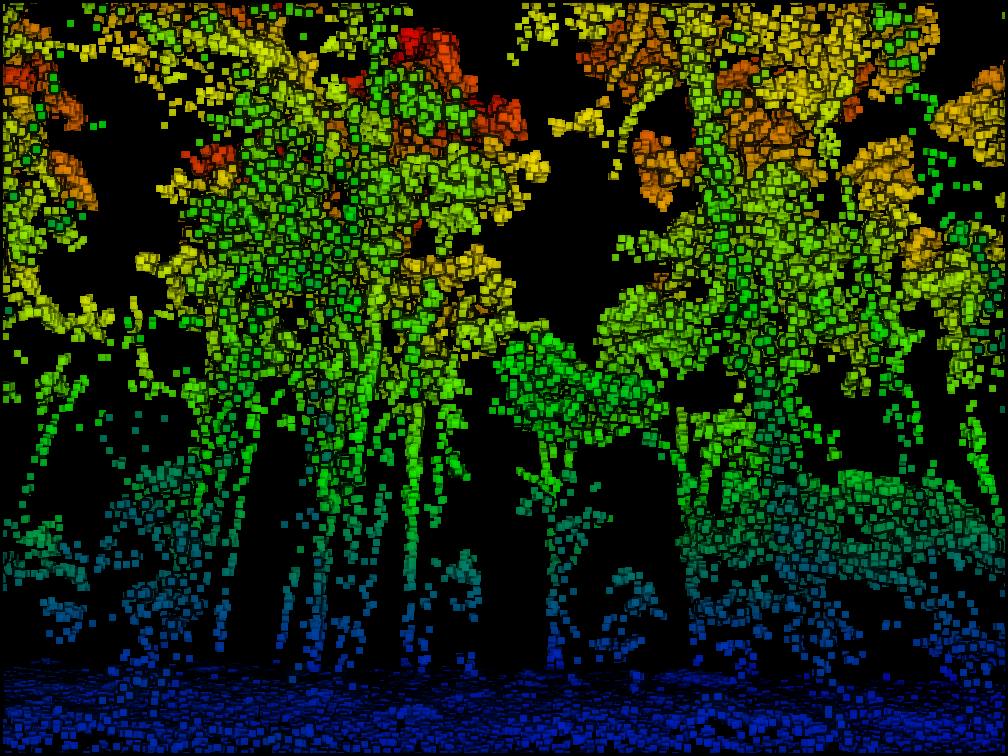} & \includegraphics[width=0.23\linewidth]{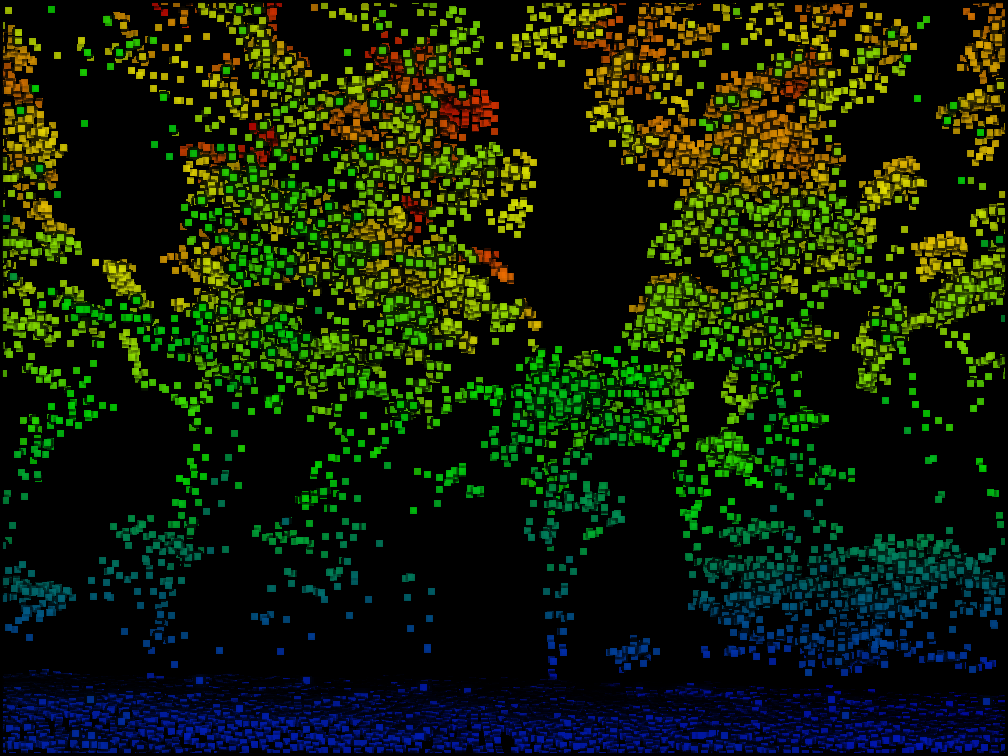} \\[5pt]
        \includegraphics[width=0.23\linewidth]{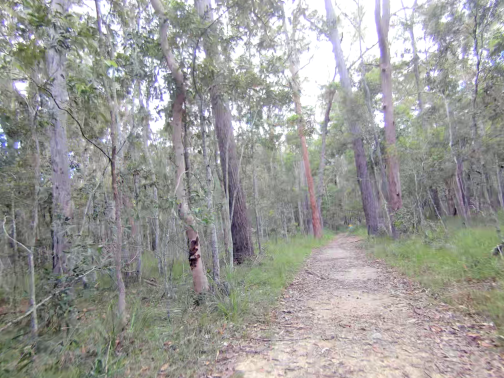} & \includegraphics[width=0.23\linewidth]{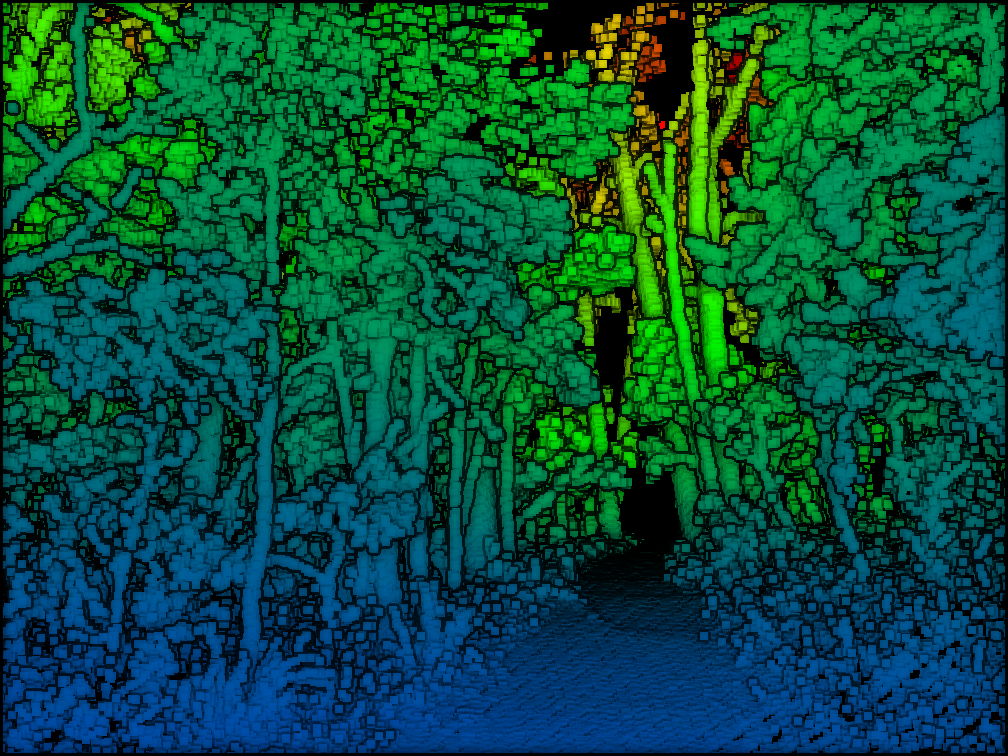} & \includegraphics[width=0.23\linewidth]{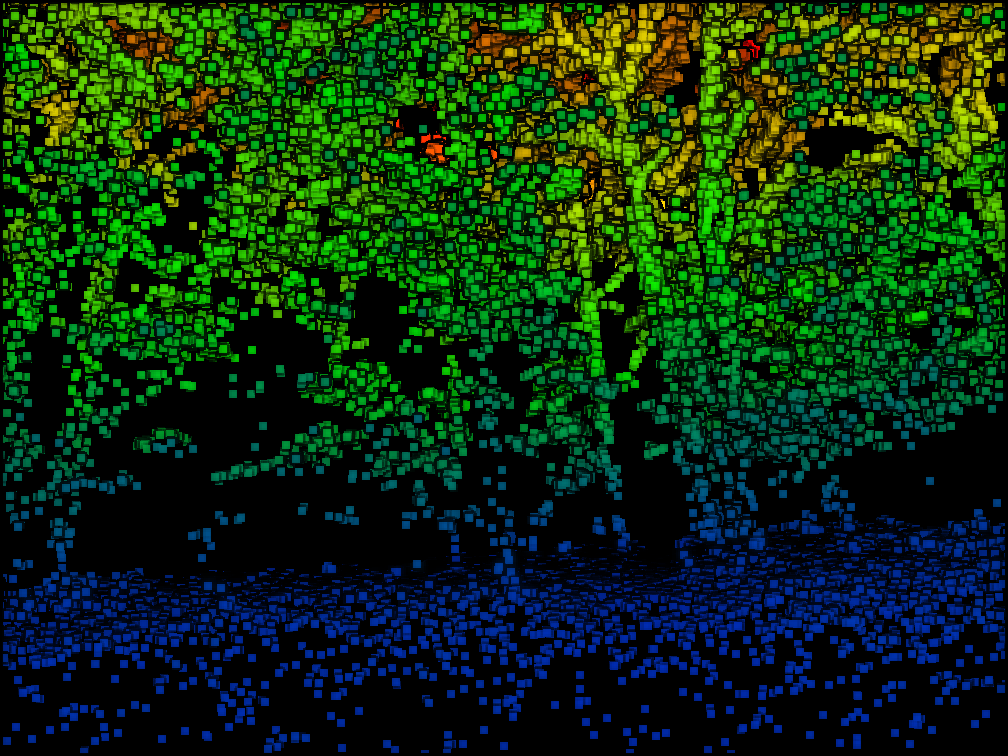} & \includegraphics[width=0.23\linewidth]{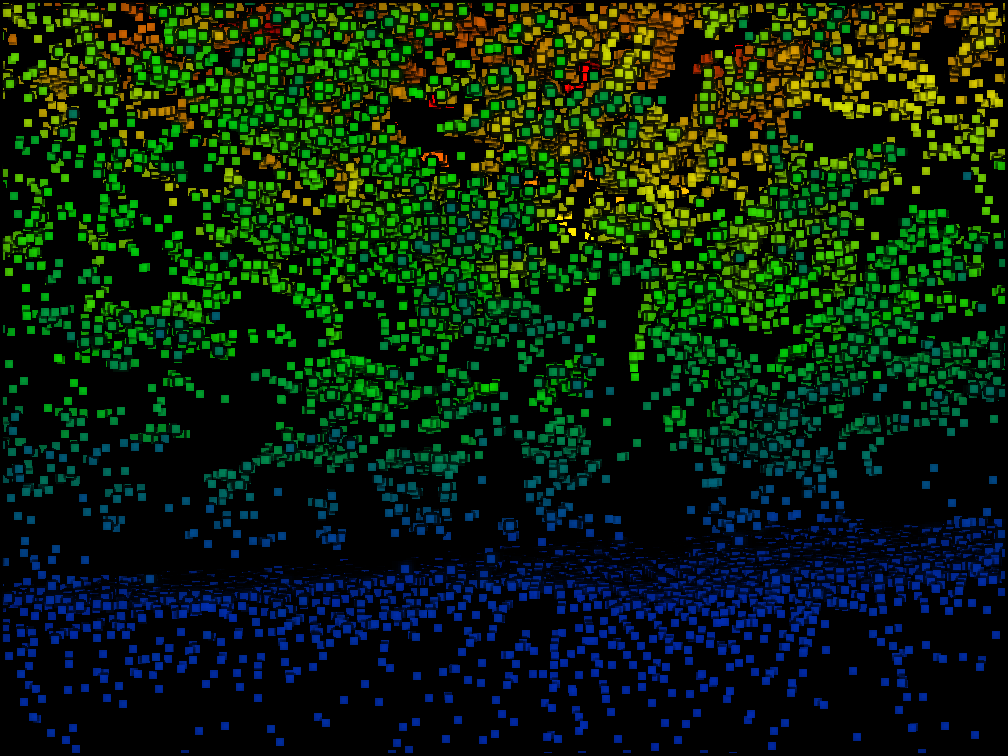} \\[5pt]
        \includegraphics[width=0.23\linewidth]{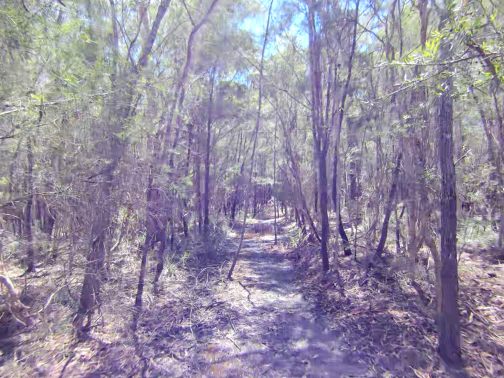} & \includegraphics[width=0.23\linewidth]{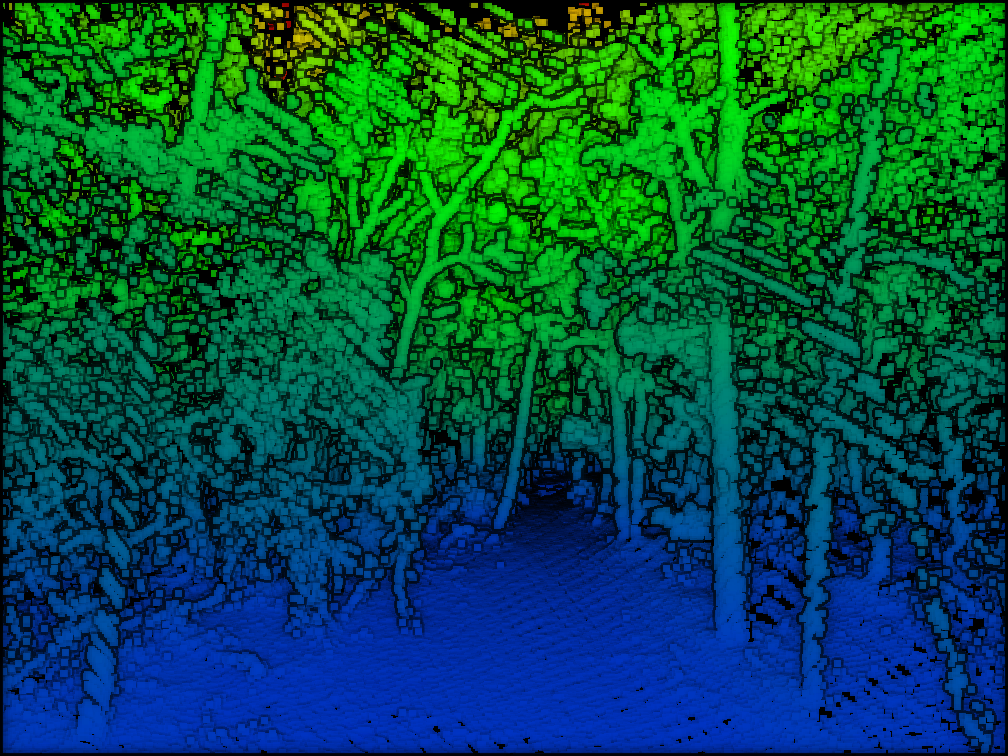} & \includegraphics[width=0.23\linewidth]{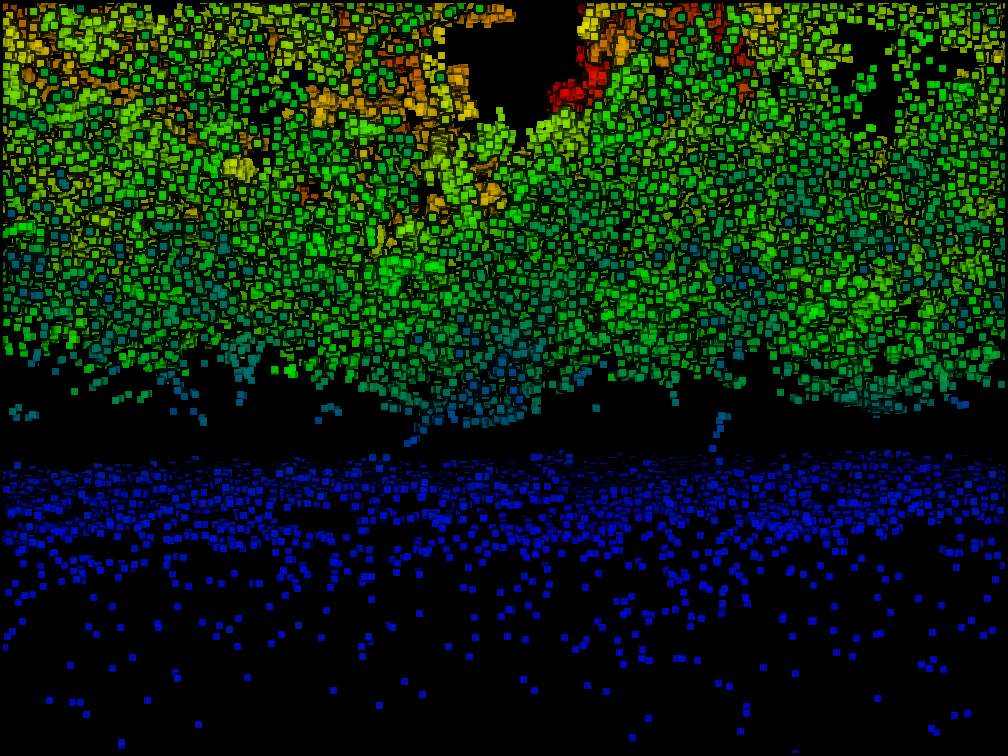} & \includegraphics[width=0.23\linewidth]{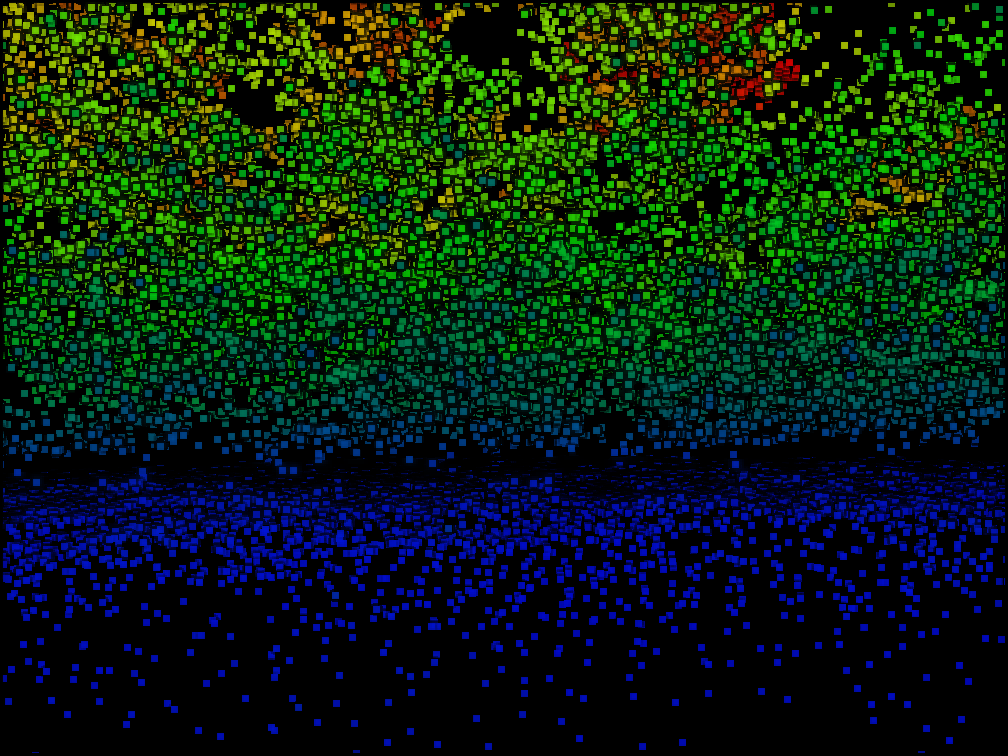} \\
        \footnotesize{(a) Ground image} & \footnotesize{(b) Ground LiDAR} & \footnotesize{(c) Low Altitude LiDAR} & \footnotesize{(d) High Altitude LiDAR} \\
    \end{tabular}
    }
    \caption{Comparison of low and high altitude aerial LiDAR submaps in \dataset{}, from the ground camera perspective. The poor beam penetration in very dense canopy exacerbates the ground--aerial matching problem.}
    \label{fig:sup_la_ha_comparison}
\end{figure*}

\Cref{fig:sup_la_ha_comparison} presents a visualisation of the low and high altitude aerial LiDAR layers of \dataset{}, in comparison to the ground LiDAR and ground camera perspective view. The effect that canopy density has on LiDAR beam penetration is evident, with significantly fewer points captured on tree trunks in the aerial LiDAR scans. HA LiDAR in particular exhibits the lowest below-canopy point density of all LiDAR types.

\section{Additional Results}
We provide additional results on \dataset{}, including Recall@N curves for the main PR experiments.

\subsection{Recall@N Curves}
\begin{figure*}[t]
    \centering
    \begin{subfigure}[b]{0.6\linewidth}
        \centering
        \includegraphics[width=\linewidth,trim={0 0 0 335},clip]{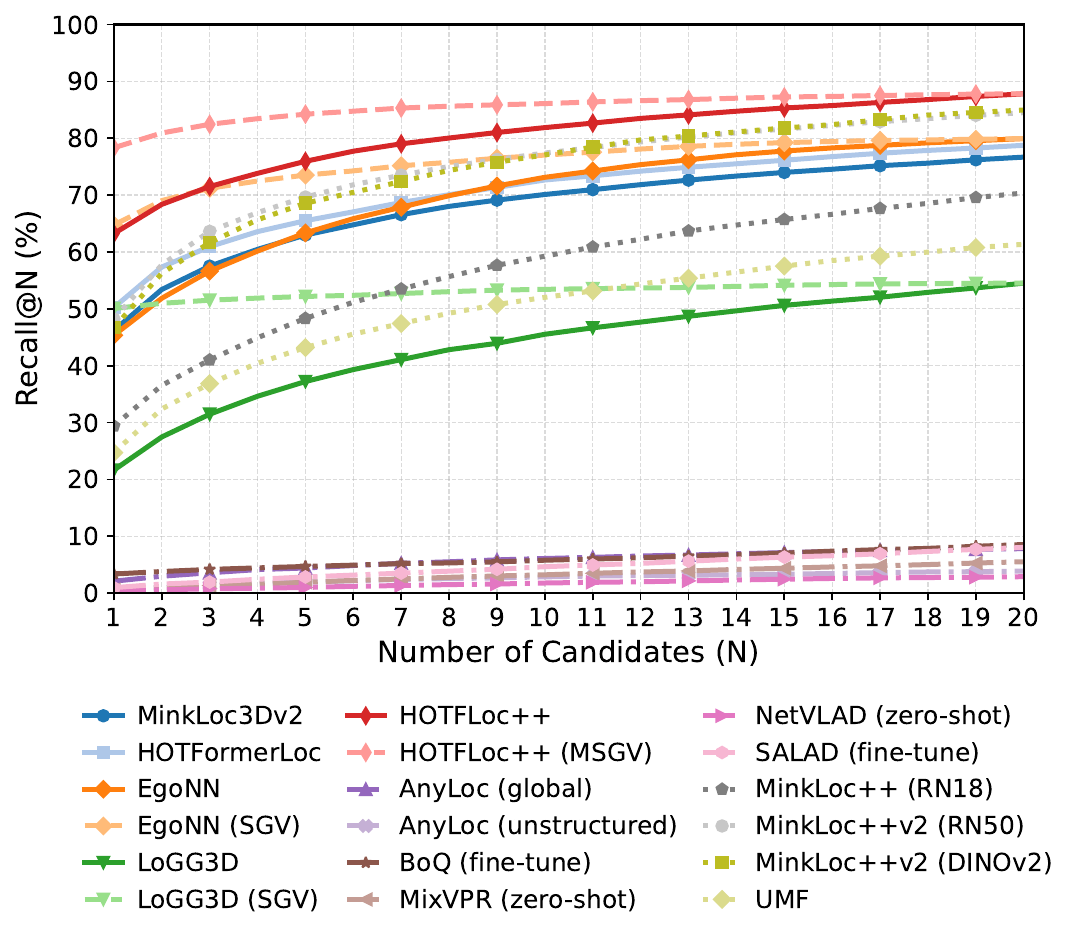}
    \end{subfigure}
    \vfill
    \begin{subfigure}[b]{0.49\linewidth}
        \centering
        \includegraphics[width=\linewidth,trim={0 120 0 0},clip]{figures/supp/recall_curves/recall_at_n_kara.pdf}
        \caption{Karawatha}
    \end{subfigure}
    \hfill
    \begin{subfigure}[b]{0.49\linewidth}
        \centering
        \includegraphics[width=\linewidth,trim={0 120 0 0},clip]{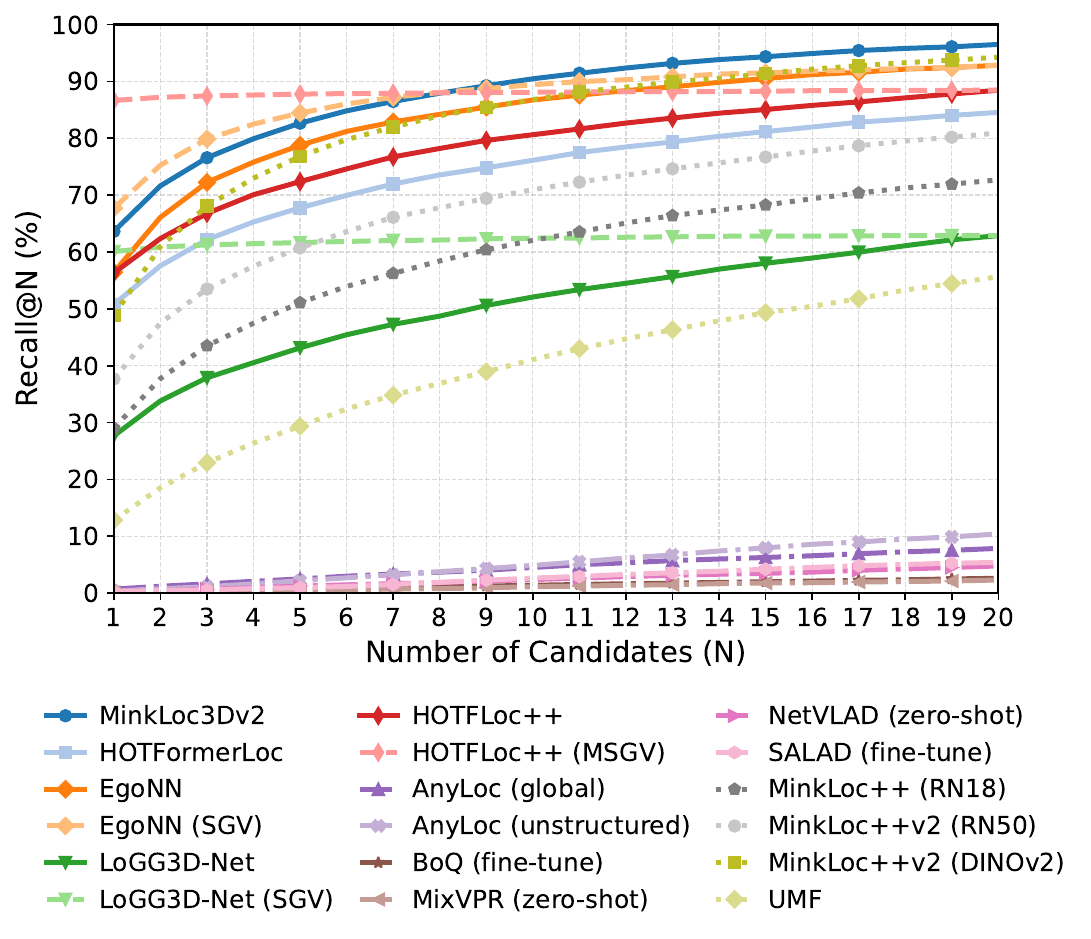}
        \caption{Venman}
    \end{subfigure}
    \vfill
    \begin{subfigure}[b]{0.49\linewidth}
        \centering
        \includegraphics[width=\linewidth,trim={0 65 0 0},clip]{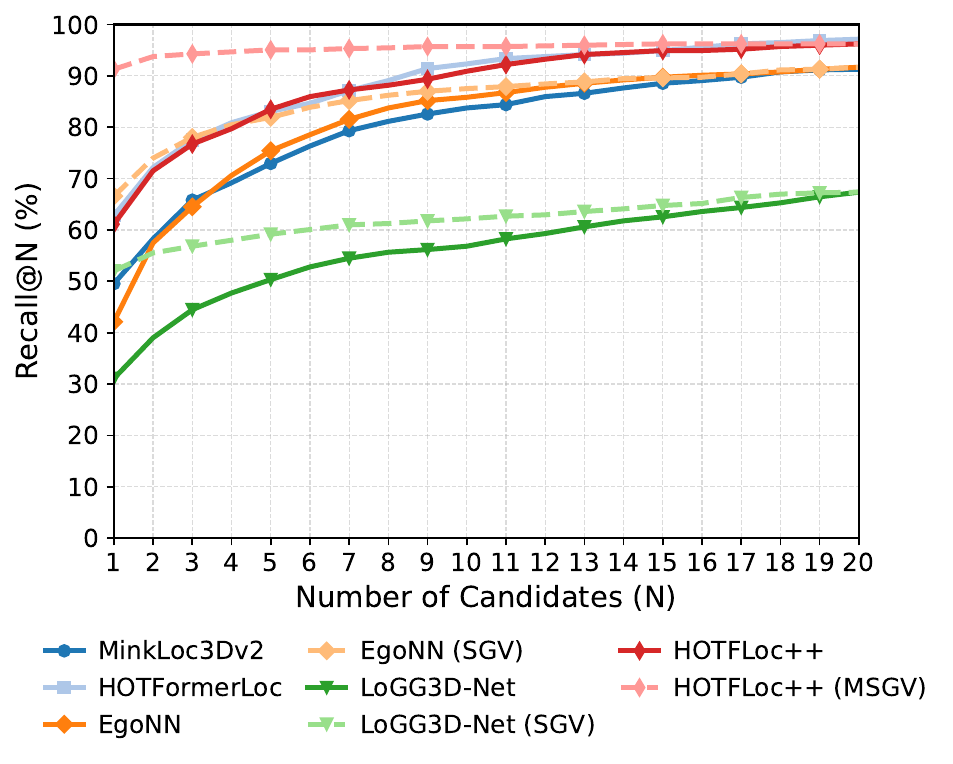}
        \caption{QCAT}
    \end{subfigure}
    \hfill
    \begin{subfigure}[b]{0.49\linewidth}
        \centering
        \includegraphics[width=\linewidth,trim={0 65 0 0},clip]{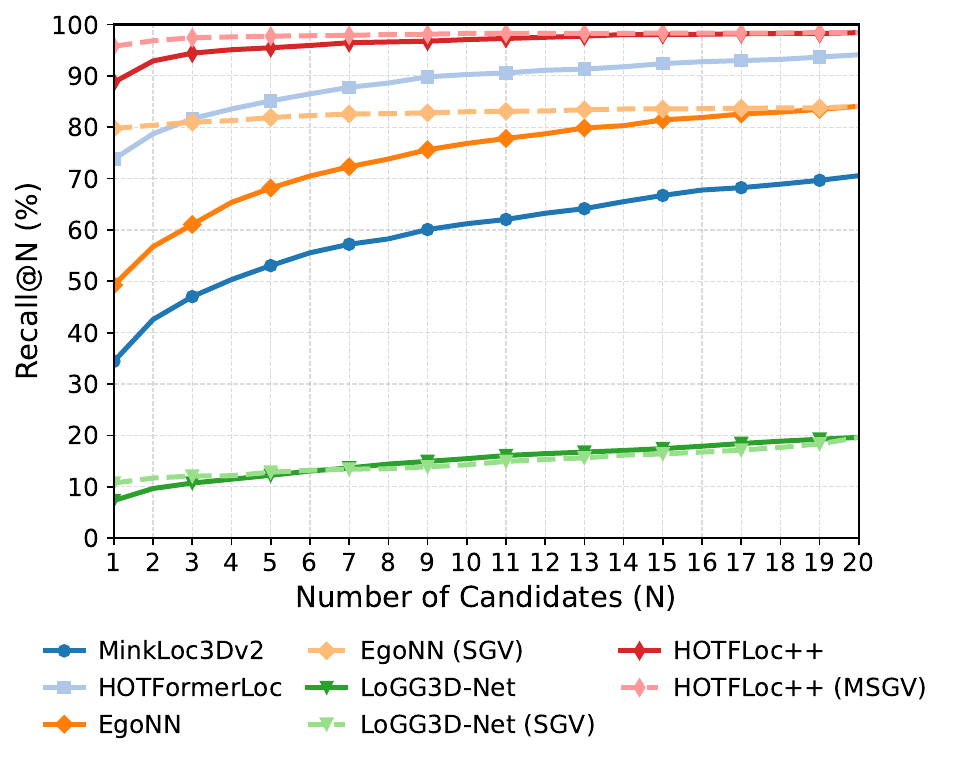}
        \caption{Samford}
    \end{subfigure}
    \caption{Average Recall@N curves for each forest in \dataset{}, comparing LPR, VPR, and MMPR models for low altitude LiDAR.}
    \label{fig:sup_recall_curves}
\end{figure*}

We compare Recall@N curves for LPR, VPR, and MMPR models on \dataset{} in \cref{fig:sup_recall_curves}, averaged over all sequences in each forest. The gap between VPR methods and LPR\,/\,MMPR is significant, highlighting that significant progress is yet to be made in applying VPR to ground--aerial PR in heavily occluded environments.

